\documentclass[10pt,twocolumn,letterpaper]{article}

\usepackage[pagenumbers]{cvpr} %

\usepackage{graphicx}
\usepackage{amsmath}
\usepackage{amssymb}
\usepackage{booktabs}

\usepackage{multirow} %
\usepackage{soul}%
\usepackage[table]{xcolor} %
\usepackage{changepage,threeparttable} %
\makeatletter
\@namedef{ver@everyshi.sty}{}
\makeatother
\usepackage{svg}
\usepackage{bm}
\usepackage[accsupp]{axessibility}  %

\newcommand{\cdf}{\Phi_s}
\newcommand{\expo}[1]{\exp\left(#1\right)}
\newcommand{\samp}{\bm{p}}
\newcommand{\opacity}{\alpha}
\newcommand{\radiance}{\bm{c}}
\newcommand{\absrp}{\sigma}
\newcommand{\irradiance}{C_{\mathrm{in}}}
\newcommand{\objo}{\bm{x}}
\newcommand{\anyo}{\bm{u}}
\newcommand{\lightd}{\bm{l}}
\newcommand{\sdf}{f}
\newcommand{\material}{g}
\newcommand{\rayo}{\bm{o}}
\newcommand{\rayd}{\bm{v}}
\newcommand{\normal}{\bm{n}}
\newcommand{\diffobjo}{\widehat{\objo}}
\newcommand{\edgew}{w}
\newcommand{\diffobjonear}{\widehat{\objo}_\mathrm{n}}
\newcommand{\objonear}{\objo_\mathrm{n}}
\newcommand{\diffobjofar}{\widehat{\objo}_\mathrm{f}}
\newcommand{\objofar}{\objo_\mathrm{f}}
\newcommand{\irradiancepix}{\widehat{C}_{\mathrm{in}}}
\newcommand{\pixel}{p}
\newcommand{\shadowpix}{I_{\mathrm{s}}}
\newcommand{\rgbpix}{I_{\mathrm{r}}}

\newcommand{\diffuse}{\bm{\rho}_d}
\newcommand{\specular}{\bm{\rho}_s}
\newcommand{\speccoeff}{\bm{y}}
\newcommand{\specbase}{D}
\newcommand{\halfvec}{\bm{h}}
\newcommand{\lighto}{\bm{q}}

\usepackage[pagebackref,breaklinks,colorlinks]{hyperref}

\usepackage[capitalize]{cleveref}
\crefname{section}{Sec.}{Secs.}
\Crefname{section}{Section}{Sections}
\Crefname{table}{Table}{Tables}
\crefname{table}{Tab.}{Tabs.}

\def\cvprPaperID{***} %
\def\confName{CVPR}
\def\confYear{2023}

\begin{document}

\title{ShadowNeuS: Neural SDF Reconstruction by Shadow Ray Supervision}

\author{
Jingwang Ling\textsuperscript{1}
\and Zhibo Wang\textsuperscript{2}
\and Feng Xu\textsuperscript{1}\thanks{Corresponding author}
\smallskip
\and
\textsuperscript{1}School of Software and BNRist, Tsinghua University\qquad 
\textsuperscript{2}SenseTime Research
}

\maketitle

\begin{abstract}
By supervising camera rays between a scene and multi-view image planes, NeRF reconstructs a neural scene representation for the task of novel view synthesis. On the other hand, shadow rays between the light source and the scene have yet to be considered. Therefore, we propose a novel shadow ray supervision scheme that optimizes both the samples along the ray and the ray location. By supervising shadow rays, we successfully reconstruct a neural SDF of the scene from single-view images under multiple lighting conditions. Given single-view binary shadows, we train a neural network to reconstruct a complete scene not limited by the camera's line of sight. By further modeling the correlation between the image colors and the shadow rays, our technique can also be effectively extended to RGB inputs. We compare our method with previous works on challenging tasks of shape reconstruction from single-view binary shadow or RGB images and observe significant improvements. The code and data are available at \url{https://github.com/gerwang/ShadowNeuS}.

\end{abstract}

\section{Introduction}
\label{sec:intro}

\begin{figure}[t]
  \centering
  \includeinkscape[width=.98\linewidth]{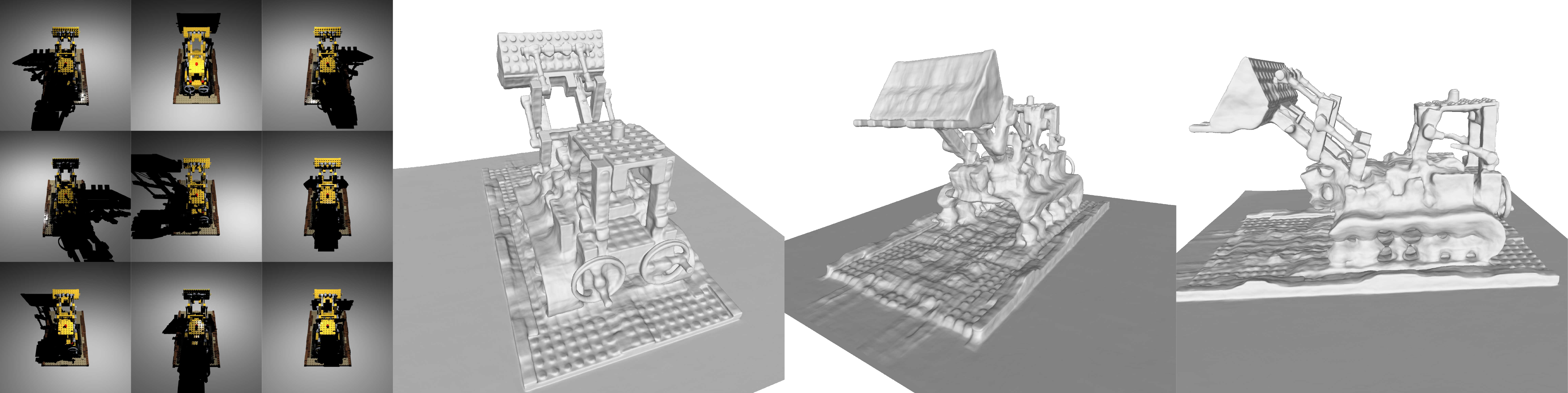}
  \includeinkscape[width=.98\linewidth]{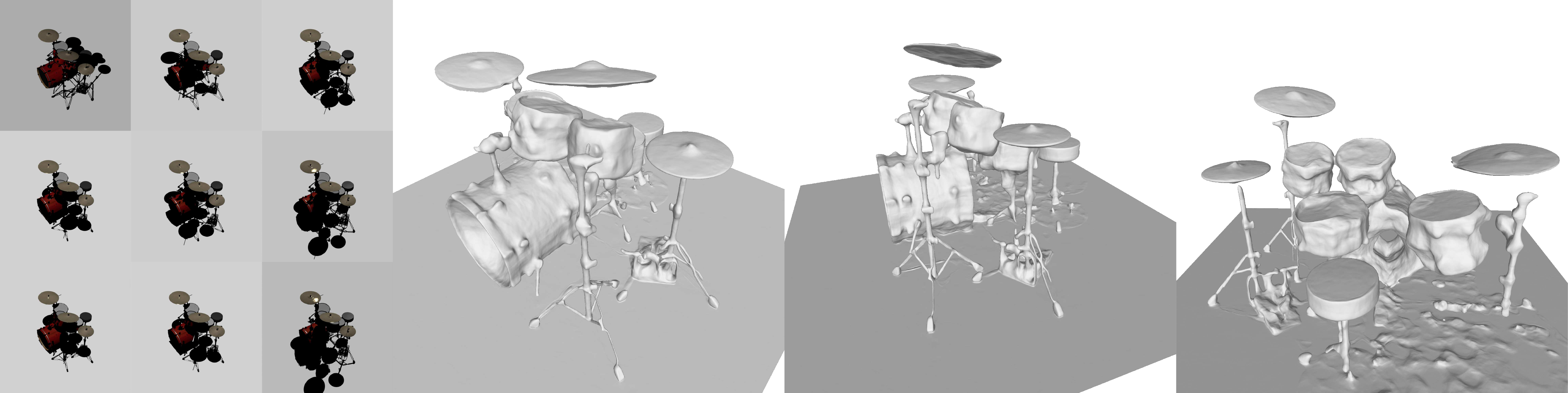}
  \includeinkscape[width=.98\linewidth]{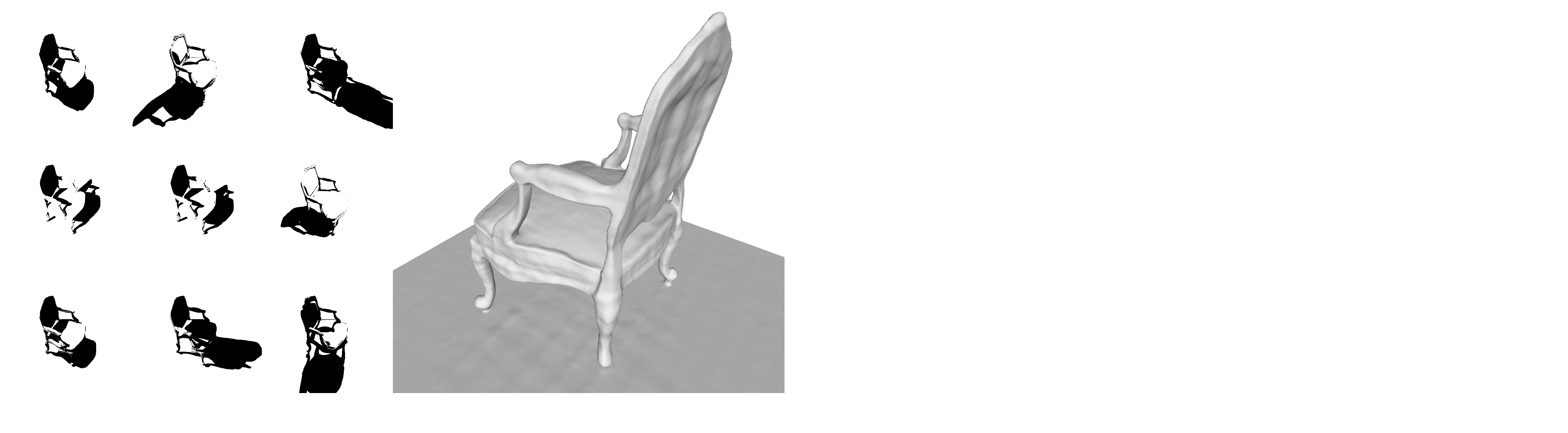}
  \caption{Our method can reconstruct neural scenes from single-view images captured under multiple lightings by effectively leveraging  a novel shadow ray supervision scheme.}
  \label{fig:teaser}
\end{figure}

Neural field \cite{DBLP:journals/cgf/XieTSLYKTTSS22} has been used for 3D scene representation in recent years.
It achieves remarkable quality because of the ability to continuously parameterize a scene with a compact neural network.
The neural network nature makes it amenable to various optimization tasks in 3D vision, including long-standing problems like image-based\cite{DBLP:conf/cvpr/NiemeyerMOG20,DBLP:conf/nips/YarivKMGABL20} and point cloud-based\cite{DBLP:conf/cvpr/ParkFSNL19,DBLP:conf/cvpr/MeschederONNG19} 3D reconstruction.
So more and more works are using neural fields as the 3D scene representation for various related tasks.

Among these works, NeRF\cite{DBLP:conf/eccv/MildenhallSTBRN20} is a representative method that incorporates a part of physically based light transport\cite{DBLP:journals/tog/SteinbergY21} into the neural field.
The light transport describes light travels from the light source to the scene and then from the scene to the camera. 
NeRF considers the latter part to model the interaction between the scene and the cameras along the {\it camera rays} (rays from the camera through the scene).
By supervising these camera rays of different viewpoints with the corresponding recorded images, NeRF optimizes a neural field to represent the scene.
Then NeRF casts camera rays from novel viewpoints through the optimized neural field to generate novel-view images.

However, NeRF does not model the rays from the scene to the light source, which motivates us to consider: can we optimize a neural field by supervising these rays?
These rays are often called {\it shadow rays} as the light emitted from the light source can be absorbed by scene particles along the rays, resulting in varying light visibility (a.k.a. shadows) at the scene surface.
By recording the incoming radiance at the surface, we should be able to supervise the shadow rays to infer the scene geometry.

Given this observation, we derive a novel problem of supervising the shadow rays to optimize a neural field representing the scene, analogizing to NeRF that models the camera rays.
Like multiple viewpoints in NeRF, we illuminate the scene multiple times using different light directions to obtain sufficient observations.
For each illumination, we use a fixed camera to record the light visibility at the scene surface as supervision labels for the shadow rays.
As rays connecting the scene and the light source march through the 3D space, we can reconstruct a complete 3D shape not constrained by the camera's line of sight.

We solve several challenges when supervising the shadow rays using camera inputs.
In NeRF, each ray's position can be uniquely determined by the known camera center, but shadow rays need to be determined by the scene surface, which is not given and has yet to be reconstructed.
We solve this using an iterative updating strategy, where we sample shadow rays starting at the current surface estimation. 
More importantly, we make the sampled locations differentiable to the geometry representation, thus can optimize the starting positions of shadow rays.
However, this technique is insufficient to derive correct gradients at surface boundaries with abrupt depth changes, which coincides with recent findings in differentiable rendering\cite{DBLP:journals/tog/LiADL18,DBLP:conf/cvpr/ZhangLLS22,DBLP:journals/tog/LaineHKSLA20,DBLP:journals/tog/ViciniSJ22,DBLP:conf/siggrapha/BangaruGLLSHBXB22}.
Thus, we compute surface boundaries by aggregating shadow rays starting at multiple depth candidates.
It remains efficient as boundaries only occupy a small amount of surface, while it significantly improves the surface reconstruction quality.
In addition, RGB values recorded by the camera encode the outgoing radiance at the surface instead of the incoming radiance.
The outgoing radiance is a coupling effect of light, material, and surface orientation.
We propose to model the material and surface orientation to decompose the incoming radiance from RGB inputs to achieve reconstruction without needing shadow segmentation (Row 1 and 2 in \cref{fig:teaser}).
As material modeling is optional, our framework can also take binary shadow images \cite{DBLP:journals/corr/abs-2203-15065} to achieve shape reconstruction (Row 3 in \cref{fig:teaser}).

We compare our method with previous single-view reconstruction methods (including shadow-only and RGB-based) and observe significant improvements in shape reconstruction.
Theoretically, our method handles a dual problem of NeRF. 
So, comparing the corresponding parts of the two techniques can inspire readers to get a deeper understanding of the essence of neural scene representation to a certain extent, as well as the relationship between them.

Our contributions are:
\begin{itemize}
  \item A framework that exploits light visibility to reconstruct neural SDF from shadow or RGB images under multiple light conditions.
  \item A shadow ray supervision scheme that embraces differentiable light visibility by simulating physical interactions along shadow rays, with efficient handling of surface boundaries.
  \item Comparisons with previous works on either RGB or binary shadow inputs to verify the accuracy and completeness of the reconstructed scene representation.
\end{itemize}

\section{Related Work}
\label{sec:related_work}

\noindent {\bf Neural fields for 3D reconstruction.}
A neural field\cite{DBLP:journals/cgf/XieTSLYKTTSS22} typically parameterizes a 3D scene with a multi-layer perceptron (MLP) network that takes scene coordinates as input.
It can be supervised with 3D constraints like point clouds\cite{DBLP:conf/cvpr/ParkFSNL19,DBLP:conf/cvpr/MeschederONNG19} to reconstruct an implicit representation of 3D shapes.
It is also possible to optimize a neural field from multi-view images by differentiable rendering\cite{DBLP:conf/cvpr/NiemeyerMOG20,DBLP:conf/nips/YarivKMGABL20,DBLP:conf/siggrapha/BangaruGLLSHBXB22}.
NeRF\cite{DBLP:conf/eccv/MildenhallSTBRN20} demonstrates remarkable novel-view synthesis quality on scenes with complex geometry.
However, the density representation in NeRF is not convenient for regularizing and extracting scene surfaces.
Thus, \cite{DBLP:conf/nips/WangLLTKW21,DBLP:conf/nips/YarivGKL21,DBLP:conf/iccv/OechsleP021} propose to combine NeRF with surface representation to reconstruct high-quality and well-defined surfaces.
While all the above works require known camera viewpoints, \cite{DBLP:journals/corr/abs-2102-07064,DBLP:conf/iccv/LinM0L21,DBLP:journals/corr/abs-2204-05735} explore to optimize camera parameters with the neural field jointly.

NeRF does not model the light source and assumes the scene emits the light. 
This assumption is suitable for view synthesis but not relighting.
Several works extend NeRF to relighting, where shadows are an essential factor.
\cite{DBLP:conf/eccv/BiXSHHKR20,DBLP:journals/corr/abs-2008-03824,DBLP:conf/cvpr/ZhangLLS22} require co-located camera-light setup to avoid shadows in captured images.
\cite{DBLP:conf/cvpr/ZhangLWBS21,DBLP:conf/iccv/BossBJBLL21,DBLP:conf/nips/BossJBLBL21} assume smooth environment lights and ignore shadows.
\cite{DBLP:conf/cvpr/SrinivasanDZTMB21,DBLP:journals/corr/abs-2112-05140,DBLP:conf/cvpr/ZhangSHFJZ22,DBLP:conf/eccv/YaoZLQFMTQ22,DBLP:conf/eccv/YangCCCW22,DBLP:conf/eccv/ChenL22} adopt neural networks conditioned on the light direction to model light-dependent shadows.
Among them, \cite{DBLP:conf/cvpr/ZhangSHFJZ22,DBLP:journals/tog/ZhangSDDFB21,DBLP:conf/eccv/YaoZLQFMTQ22,DBLP:conf/eccv/YangCCCW22,DBLP:conf/eccv/ChenL22} first reconstruct geometry using multi-view stereo and compute shadows using fixed geometry.
None of the works refine the geometry to match the shadows in the captured images.
However, we show that it is possible to reconstruct a complete 3D shape from scratch by exploiting information in the shadows.

\noindent {\bf Single-view reconstruction.}
\cite{DBLP:conf/cvpr/YuYTK21,DBLP:conf/iccv/JainTA21,DBLP:conf/eccv/XuJWFSW22} explore reconstructing neural fields from a few or a single image, but they require data-driven prior in the pretrained networks thus are in a different scope from ours.
Non-line-of-sight imaging\cite{DBLP:journals/nature/OTooleLW18,DBLP:conf/cvpr/XinNKSNG19,DBLP:journals/pami/ShenWLPLGLY21} adopts a transient sensor to capture time-resolve signals, which enables reconstructing the scene beyond the camera's view frustum.
Photometric stereo\cite{DBLP:conf/cvpr/LiL22,DBLP:conf/cvpr/ChenHSMW19} reconstructs surface normals from images captured under directional lights.
Normals can be integrated to produce a depth map but require non-trivial processing\cite{DBLP:conf/cvpr/CaoSOM21,DBLP:conf/eccv/CaoSSOM22}.

\noindent {\bf Shape from Shadows.}
Shadows indicate varying incoming radiance caused by occlusion, providing scene geometry cues.
There is a long history of reconstructing shapes from shadows as 1D curves\cite{DBLP:conf/aaai/KenderS86,DBLP:conf/cvpr/HatzitheodorouK88}, 2D height maps\cite{DBLP:journals/trob/RavivPL89,DBLP:conf/cvpr/DaumD98,DBLP:journals/ijcv/YuC05,DBLP:conf/iccv/SavareseRBP01} and 3D voxel grids\cite{DBLP:conf/iros/LangerDZ95,DBLP:journals/ijcv/SavareseARBP07,DBLP:journals/ijcv/YamazakiNBK09}.
These works typically capture under different light directions to get sufficient observations of shadows.
Shadows show the potential in these works to reconstruct surface details\cite{DBLP:journals/ijcv/YuC05} and intricate thin structures\cite{DBLP:journals/ijcv/YamazakiNBK09}.
The most recent work in this area is DeepShadow\cite{DBLP:journals/corr/abs-2203-15065}, which reconstructs a neural depth map from shadows.
A different setup with fixed lighting but multiple viewpoints is also adopted by \cite{DBLP:journals/corr/abs-2203-15946}, which integrates Shadow Mapping to reconstruct a neural representation.
Concurrently and independently, \cite{DBLP:journals/corr/abs-2210-08936} proposes to simultaneously use shading and shadows in neural field reconstruction.
In particular, they compute shadows at a {\it non-differentiable} surface point located by root finding, making it rely on a differentiable shading computation.
We propose fully differentiable shadow ray supervision that optimizes both the shadow ray samples and the surface point, enabling neural field reconstruction from either pure shadows or RGB images.

\section{Ray Supervision in Neural Fields}

This section first reveals the essence in NeRF\cite{DBLP:conf/eccv/MildenhallSTBRN20} training as supervising {\it camera rays}.
From there, we discover a ray supervision scheme generalizable to arbitrary rays.
The scheme makes it feasible for {\it shadow rays} to supervise the optimization of a neural scene representation.

\begin{figure}[t]
  \centering
  \includegraphics[width=\linewidth]{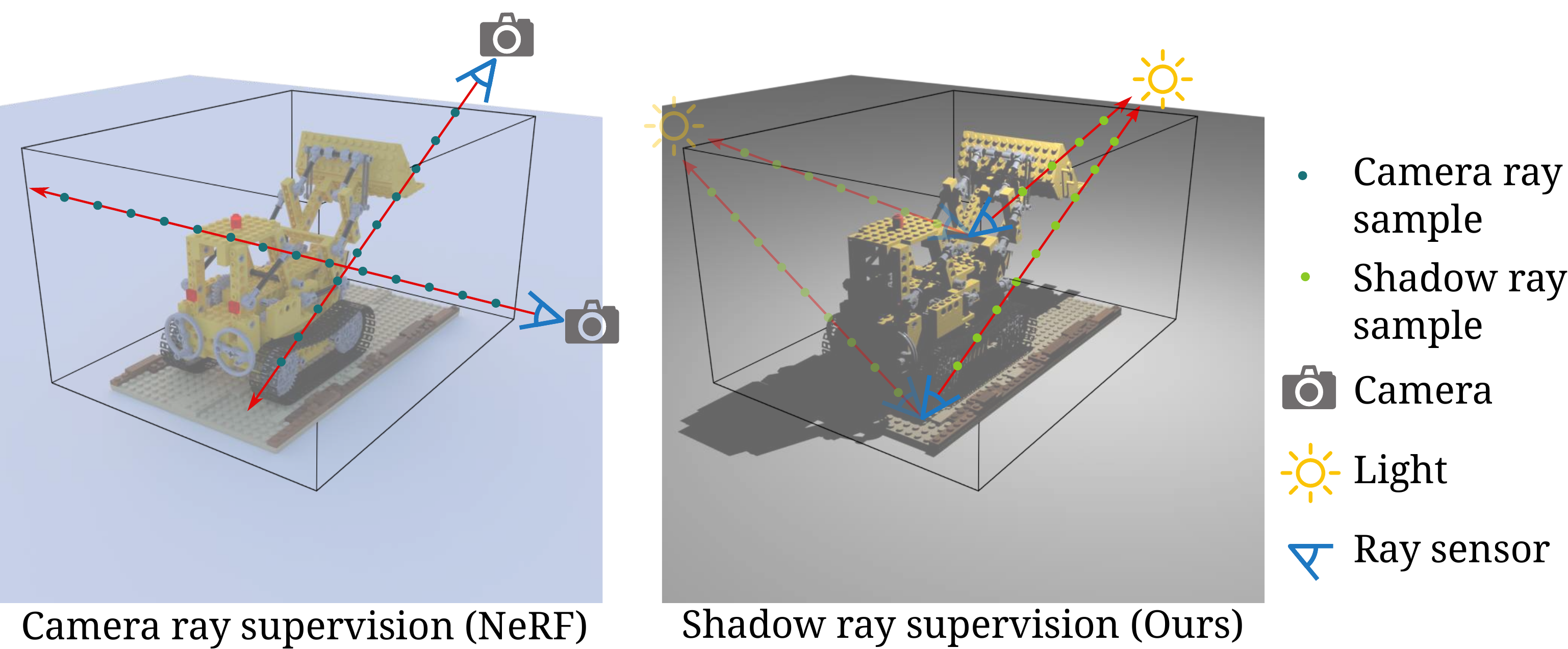}
  \caption{Different kinds of ray supervisions.}
\vspace{-1em}
  \label{fig:ray_supervision}
\end{figure}

\subsection{Camera ray supervision in NeRF}

NeRF aims to optimize a neural field to fit a scene of interest.
To obtain observations of the scene, NeRF requires recording images at multiple camera viewpoints with known camera parameters.
Each image pixel records the incoming radiance of a camera ray that passes through the known camera center from a known direction.
Since NeRF does not model the external light source and assumes the light is emitted from scene particles to simplify the modeling of a scene with fixed lighting, the incoming radiance is actually attributed to the combined effect of light absorption and emission by the infinitesimal particles along the camera ray.%
To fit observations, NeRF uses differentiable volume rendering to simulate the same camera ray in the neural field.
NeRF uses quadrature to approximate the continuous integral in volume rendering by sampling $N$ distances $t_1, \cdots, t_N$, started from the camera center $\rayo$ along the camera ray direction $\rayd$. 
With the scene density $\absrp_i$ and emitted radiance $\radiance_i$ at each sample point $\samp(t_i)=\rayo + t_i\rayd$, the estimated radiance $C$ at the camera can be formulated as follows,
\begin{equation}
  \label{eq:nerf_render}
  C(\rayo, \rayd) = \sum_{i = 1}^{N} T_i \opacity_i \radiance_i,
\end{equation}
where $\alpha_i=1-\expo{-\absrp_i (t_{i+1} - t_i)}$ is the discrete opacity and $T_i = \exp(- \sum_{j=1}^{i-1} \absrp_j \cdot (t_{j+1} - t_j))$ indicates the light transmittance, \ie, the proportion of the emitted light reach the camera from the point $\mathbf{p}(t_i)$.
The incoming radiance recorded at the pixel can be used to supervise the simulated radiance $C$.
NeRF trains on a random subset of camera rays in each iteration.
As the neural field receives supervision signals from many camera rays marching in different viewpoint directions, it obtains sufficient scene information to optimize the neural field in the space these rays go through.

\subsection{Generalized ray supervision}

The reason that NeRF can supervise the camera rays to optimize a neural field is that multi-view cameras record the radiance as labels of the rays. 
Moreover, as each camera is calibrated, each recorded ray's 3D location and orientation are well-defined.
We can regard each pixel of the multi-view camera as a ``ray sensor'' recording the incoming radiance of a particular ray because each pixel is used independently in training.
These ray sensors are the key to the NeRF techniques.
More generally, if we let the ``ray sensors'' record other kinds of rays in the scene, it is also possible to achieve scene reconstruction.
This motivates us to consider whether we can supervise other rays and design ray sensors to record their radiance.

\subsection{Shadow ray supervision}

\begin{figure*}[t]
  \centering
  \includegraphics[width=\linewidth]{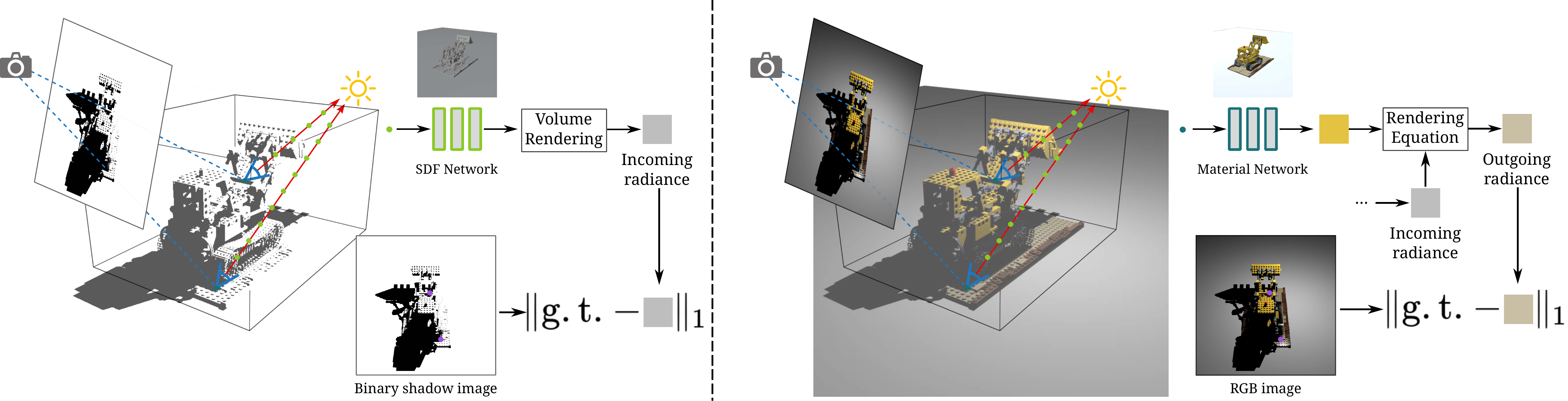}
  \caption{Overview of our method. The proposed shadow ray supervision can be applied to single-view neural scene reconstruction on two input types: binary shadow images (left) and RGB images (right). For binary inputs, we first compute the incoming radiance of a shadow ray using volume rendering. Then, we construct a photometric loss to train the neural SDF to match the shadows. For RGB inputs, we further use a material network and a rendering equation to convert the incoming radiance to the outgoing radiance. The SDF and material networks are trained to match the ground truth colors.}
\vspace{-1em}
  \label{fig:pipeline}
\end{figure*}

Since camera rays have achieved great success in neural scene reconstruction, as the counterpart in light transport, the ray connecting the scene and the light source, a.k.a. {\it shadow rays}, should also be able to be used to reconstruct neural scenes.
We first consider an ideal setup where many hypothetical ray sensors are placed in the scene at different but known locations, as shown in \cref{fig:ray_supervision}.%
To observe the scene along shadow rays, we illuminate the scene with a known directional light.
Each ray sensor captures one ray that passes the sensor from the light direction.
Different from NeRF, as we model the light source, we assume the scene does not emit light, which is more physically correct and can simplify the following process.
Therefore, the incoming radiance at a ray sensor is from the light emitted from the light source and absorbed by infinitesimal particles along the ray.
Using similar quadrature as \cref{eq:nerf_render}, we can express the incoming radiance simulated in the neural field as
\begin{equation}
  \label{eq:shadow_ray_render}
  \irradiance(\objo, \lightd) =  L\prod_{i=1}^{N}(1-\opacity_i),
\end{equation}
where and $L$ is the intensity of the light source, $\objo$ is the location of a ray sensor and $\lightd$ is the light direction.
To obtain sufficient information to constrain the optimization, we require the shadow rays to march the scene in different directions.
Therefore, we illuminate the scene with multiple light directions one by one and record the incoming radiance each time.
As this ray supervision scheme has been demonstrated successful by NeRF, it is also promising to reconstruct a neural scene here.
\section{Shadow ray supervision with a single-view camera}

Note that in the above formulation, we adopt hypothetical ray sensors to record the incoming radiance in the light direction and at the known positions in the scene.
These ray sensors are ideal because they are placed at desired positions in the scene and always face toward the light.
Under these strong assumptions, it is possible to get sufficient supervision for the shadow rays.
However, these ray sensors are hard to implement in an actual setup, unlike NeRF, where the ray sensors are just the pixels of multi-view cameras.
In this section, we will propose a more practical setting for a real capture setup.

In general, we conduct shadow ray supervision from a single-view camera, which can be a practical alternative to the ray sensors in the previous formulation.
We similarly illuminate the scene with a light in direction $\lightd$.
The scene is assumed to be opaque, and thus the camera captures exactly the outgoing radiance at visible surfaces.
We consider two types of camera inputs: binary shadow images\cite{DBLP:journals/corr/abs-2203-15065} and RGB images, as shown in \cref{fig:pipeline}.
Binary shadow images use outgoing radiance to determine whether a point is illuminated, which can be seen as an approximation of binarized incoming radiance.
RGB images are a more complex case that records a combined effect of material, surface orientation, and incoming radiance.
We will first consider the more straightforward case when we can obtain the incoming radiance at visible surfaces from binary shadow images and then handle the more complex RGB images.

However, another challenge is that, given the recorded pixel values, we still do not know the exact depths of the visible surface points.
Thus, we are given scene observations as outgoing radiance in the camera viewing direction at points at unknown depths.
This problem is handled by the proposed techniques that determine the depth and relate outgoing radiance to incoming radiance.

We represent the scene as the zero level set of a signed distance function (SDF) $\mathcal{S} = \left\{ \anyo \in \mathbb{R}^3 | \sdf(\anyo) = 0 \right\}$, where $\sdf$ is a neural network that regresses the signed distance at the input 3D position.
The 3D points visible by the camera are the first intersections between the camera rays and the SDF.
Note that here the camera rays are only used to determine the surface points but not to construct supervision, which is the job of shadow rays.
Specifically, ray marching\cite{DBLP:conf/nips/YarivKMGABL20} is used to compute the intersection point $\objo$ at the current SDF.
Then we can compute the incoming radiance $\irradiance(\objo,\lightd)$ at the intersection by volume rendering.
As we are modeling an SDF instead of a density field, we replace the discrete opacity $\opacity_i$ in \cref{eq:shadow_ray_render} by the one derived from the SDF following NeuS\cite{DBLP:conf/nips/WangLLTKW21}, as
\begin{equation}
  \label{eq:neus_alpha}
  \opacity_i = \max \left (1 - \frac{\cdf(\sdf(\samp(t_{i+1})))}{\cdf(\sdf(\samp(t_i)))}, 0 \right),
\end{equation}
where $\cdf(x) = (1 + e^{-sx})^{-1}$ is the sigmoid function and $s$ is a learnable scalar parameter that controls whether \cref{eq:shadow_ray_render} approaches volume rendering or surface rendering.

\noindent {\bf Differentiable intersection points.}
To locate the intersection point $\objo$ given the SDF, ray marching is the most straightforward choice.
However, as it is non-differentiable, it is prone to be misled by surface points with incorrect depths, leading to worse results.
To optimize the intersection points using backpropagated gradients, we use implicit differentiation\cite{DBLP:conf/nips/AtzmonHYIML19,DBLP:conf/nips/YarivKMGABL20}, which makes the intersection point differentiable to the SDF network parameters as
\begin{equation}\label{eq:diff_intersect}
  \diffobjo = \objo - \frac{\rayd}{\normal \cdot \rayd} \sdf(\objo),
 \end{equation}
where $\rayd$ is the camera ray direction and $\normal=\nabla_{\objo}\sdf(\objo)$ is the surface normal derived from the SDF network.
Then, we use $\irradiance(\diffobjo,\lightd)$ as the differentiable radiance at intersection $\objo$.
As $\objo$ acts as the start position of a shadow ray, it can be optimized by gradients from \cref{eq:shadow_ray_render}.
When the computed incoming radiance $\irradiance(\diffobjo,\lightd)$  does not agree with the supervision, the SDF network can optimize both the signed distances along the shadow ray and the starting position of the ray to fit the observation.

\noindent {\bf Multiple shadow rays at boundaries.}
We observe that $\diffobjo$ in \cref{eq:diff_intersect} only differentiates along the camera direction $\rayd$.
When supervising $\irradiance(\diffobjo,\lightd)$ with the recorded images, it will cause issues at pixels corresponding to surface boundaries.
At surface boundaries, a pixel spans disconnected regions at different depths, where each region occupies a part of the pixel's area.
When $\diffobjo$ moves perpendicular to the camera direction $\rayd$, it can significantly change the computed radiance at surface boundaries by changing the area proportional to each region.
If we only sample one shadow ray started at one region, it will lead to incorrect gradients similar to the case in differentiable mesh rendering\cite{DBLP:journals/tog/LiADL18,DBLP:journals/tog/LaineHKSLA20}.

Therefore, we first obtain a pixel subset $\Omega$ corresponding to surface boundaries, and a differentiable area ratio $\edgew$ for each boundary pixel using the surface walk procedure in \cite{DBLP:conf/cvpr/ZhangLLS22}.
Then we locate two intersections $\objonear$ and $\objofar$ at different depths within the pixel and compute their incoming radiance $\irradiance(\diffobjonear,\lightd)$ and $\irradiance(\diffobjofar,\lightd)$ respectively.
When computing the incoming radiance corresponding to pixel $\pixel$, we average the incoming radiance at boundary pixels as
\begin{equation}
  \label{eq:edge_vis}
  \irradiancepix=\left\{
    \begin{aligned}
    &\irradiance(\diffobjo,\lightd) & \pixel\notin \Omega\\
    &\edgew \irradiance(\diffobjonear,\lightd) + (1-\edgew)\irradiance(\diffobjofar,\lightd) & \pixel\in \Omega
    \end{aligned}
  \right.
\end{equation}

Then, we can supervise the computed incoming radiance $\irradiancepix$ with a pixel $\shadowpix$ on a binary shadow image as
\begin{equation}
  \label{eq:loss_shadow}
  \mathcal{L}_{\mathrm{shadow}} = \|\irradiancepix - \shadowpix\|_1.
\end{equation}

\noindent {\bf Decomposing incoming radiance by inverse rendering.}
To cope with RGB images, we incorporate an inverse rendering equation consisting of material, incoming radiance, and surface orientation.
We model the non-Lambertian BRDF as a diffuse component $\diffuse$ and a specular component $\specular$.
Following \cite{DBLP:conf/cvpr/LiL22,DBLP:conf/eccv/YangCCCW22}, we use a weighted combination of spherical Gaussian basis to represent the specular component $\specular$ as $\specular=\speccoeff^{T} \specbase(\halfvec,\normal)$, where $\halfvec=\frac{\lightd-\rayd}{\|\lightd-\rayd\|}$ is the half-vector between light direction $\lightd$ and view direction $-\rayd$, $\specbase$ is the specular basis and $\speccoeff$ is the specular coefficients.
We model another MLP network $\material$ to regress material properties $(\diffuse,\speccoeff)=\material(\objo)$ at surface location $\objo$.

The outgoing radiance at point $\objo$ can be formulated as
\begin{align}
    \label{eq:render_rgb}
    C(\objo,-\rayd)=(\diffuse + \specular)\irradiance(\objo, \lightd)(\lightd\cdot\normal)
\end{align}
The outgoing radiance $\widehat{C}$ corresponding to a boundary pixel is the weighted combination of multiple samples, similar to \cref{eq:edge_vis}.
Now we can supervise the computed radiance using a pixel $\rgbpix$ on an RGB image as
\begin{equation}
  \label{eq:loss_rgb}
  \mathcal{L}_{\mathrm{rgb}} = \|\widehat{C} - \rgbpix\|_1
\end{equation}

\noindent {\bf Light source modeling.}
Our technique supports directional light or point light as the light source to compute the incoming radiance in \cref{eq:shadow_ray_render}.
For directional light, the light direction $\lightd$ and intensity $L$ are known and uniform for all shadow rays.
For point light, we calculate the light direction and intensity at point $\objo$ as
\begin{align}
  L=\frac{L_p}{\Vert \lighto-\objo\Vert_2^2},\; \lightd=\frac{\lighto-\objo}{\Vert \lighto-\objo\Vert_2} \label{eq:light_source}
\end{align}
where $L_p$ is a scalar point light intensity and $\lighto$ is the light location.

\noindent {\bf Training.}
To regularize the network to output valid SDF, we add an Eikonal loss\cite{DBLP:conf/icml/GroppYHAL20} on $M$ sample points as
\begin{equation}
    \mathcal{L}_{\mathrm{eik}} = \frac{1}{M}\sum_{i}^{M}(\|\nabla f(\samp_{i})\|_2 - 1)^2.
\end{equation}
We train the Eikonal loss with \cref{eq:loss_shadow} or \cref{eq:loss_rgb} depending on whether binary shadow images or RGB images are used as supervision.

Our technique is mainly evaluated on bounded scenes of an object on the ground.
To bound the camera rays, we set camera rays that do not intersect with the SDF to intersect with the ground.
To resolve the scale ambiguity from single-view inputs and reconstruct a scene with the accurate scale, we assume the ground plane's position and orientation are known.
More discussion on the handling of the ground plane can be found in the supplementary material.

\begin{figure*}[t]
  \centering
  \begin{minipage}[t]{.47\textwidth}
  \includeinkscape[width=.98\linewidth]{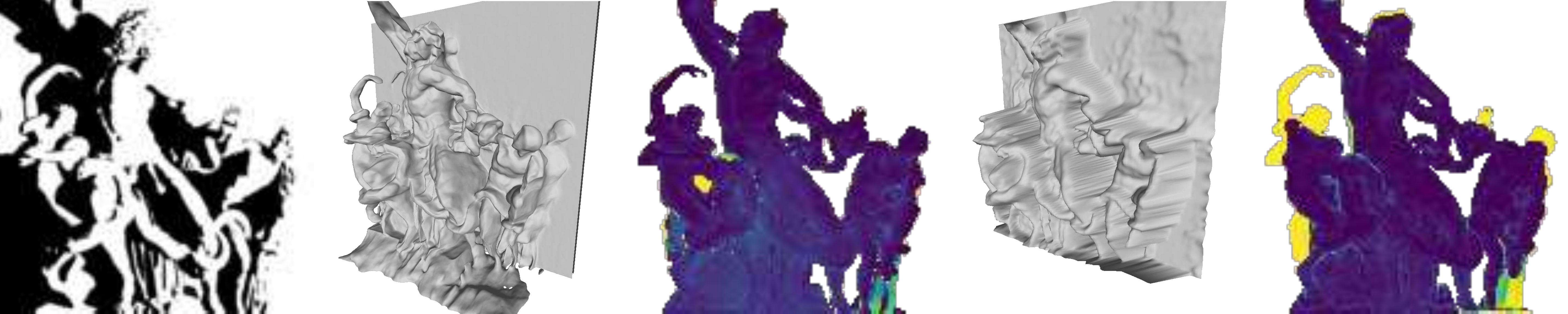}
  \includeinkscape[width=.98\linewidth]{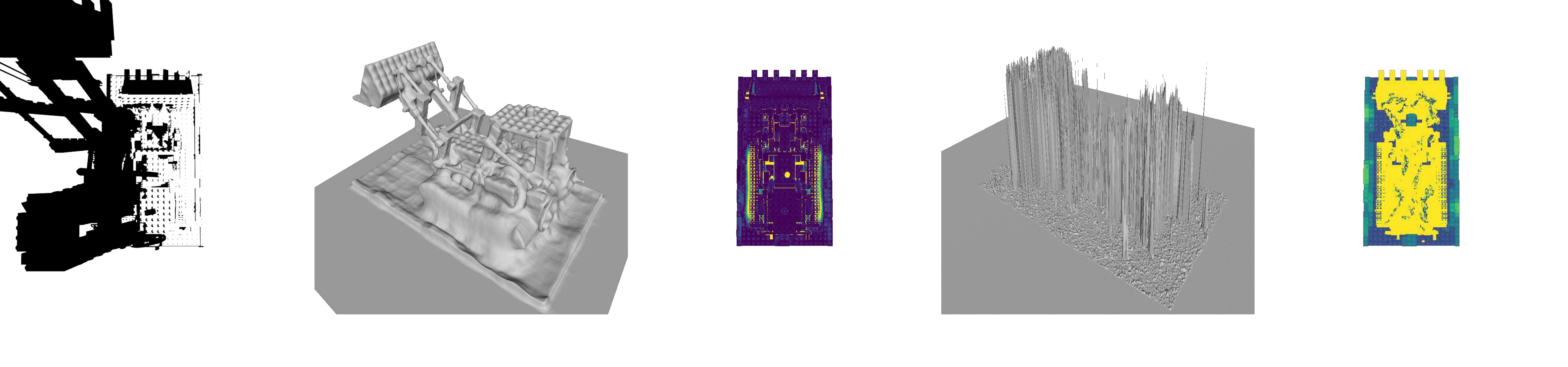}
  \end{minipage}
  \begin{minipage}[t]{.47\textwidth}
  \includeinkscape[width=.98\linewidth]{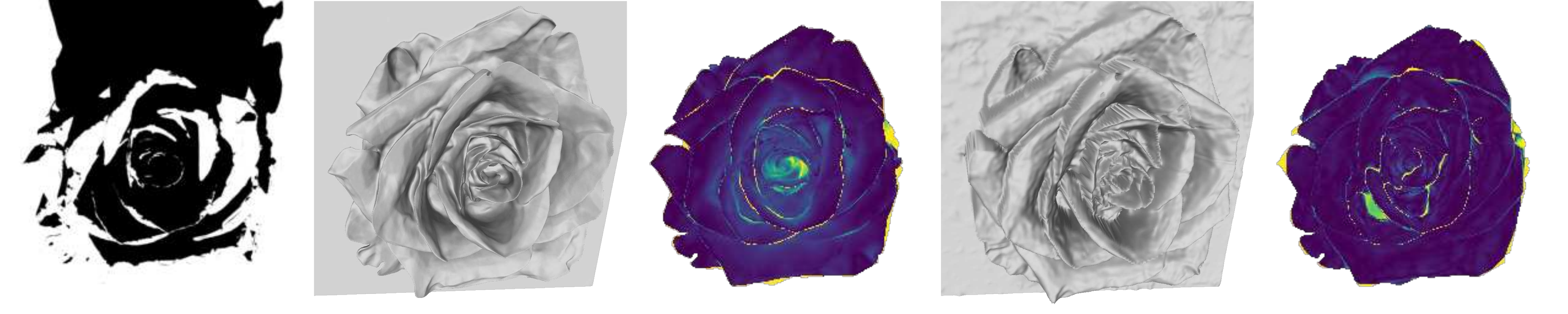}
  \includeinkscape[width=.98\linewidth]{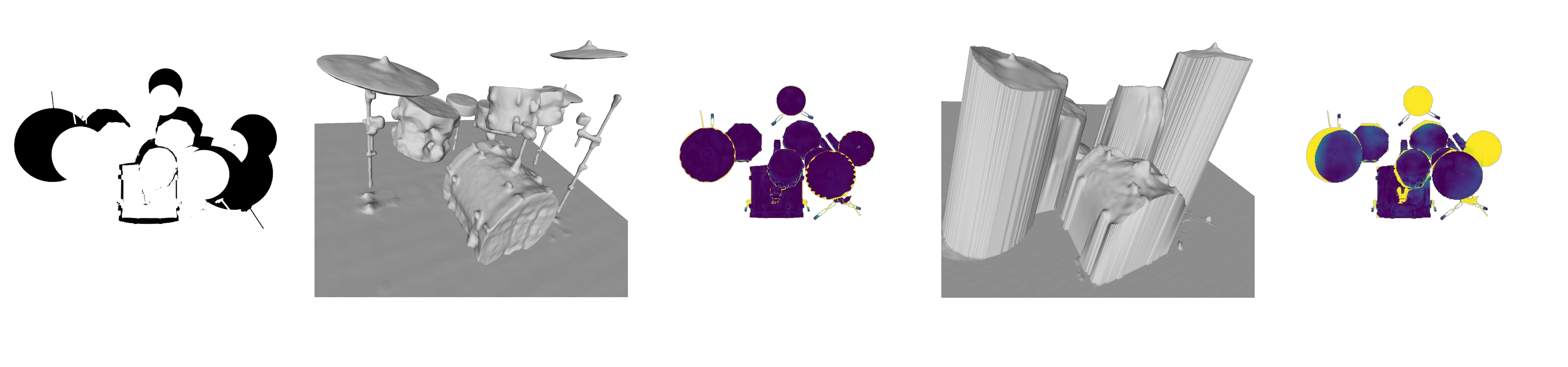}
  \end{minipage}
  \begin{minipage}[t]{.05\textwidth}
    \vspace{.00001em}
      \includeinkscape[width=.15\linewidth]{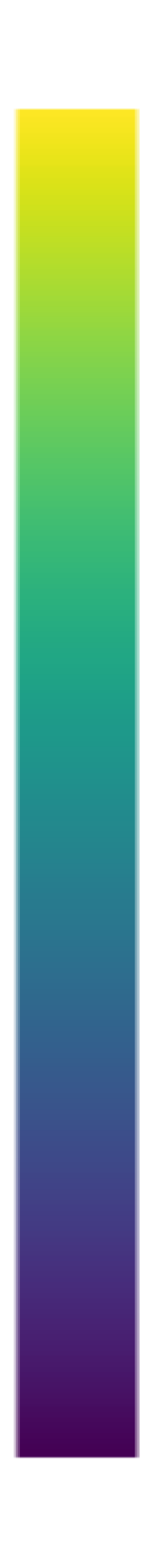}
  \end{minipage}
  \caption{Comparison on binary shadow inputs. Each result's heat map shows error distribution compared to the ground truth depth map.}
  \label{fig:shadow_quali}
  \vspace{-1em}
\end{figure*}

\section{Experiments}

\subsection{Implementation details}
We adopt an SDF MLP network similar to NeuS\cite{DBLP:conf/nips/WangLLTKW21} for both the binary shadow inputs and RGB inputs.
When handling RGB inputs, the SDF network outputs an extra 256-dimensional feature vector. 
It will be concatenated with 3D position and surface normal to regress diffuse and specular coefficients by another MLP network. 
During training, we randomly select four images in each batch, and for each image, 256 pixel positions are sampled as supervision signals.
The camera ray intersection points are located by ray marching, and possible surface boundaries are computed using a surface walk process \cite{DBLP:conf/cvpr/ZhangLLS22} started at these intersection points.
We train the network for 150k iterations, which takes about 24 hours on a single RTX 2080Ti.
More implementation details can be found in the supplementary material.

\subsection{Evaluation}

To demonstrate the ability to leverage information from shadow rays in scene reconstruction, we evaluate our method on single-view binary shadow images and RGB images captured under multiple known light directions. We first present qualitative and quantitative comparisons with state-of-the-art methods supporting similar inputs. Then, we evaluate the effectiveness of the shadow ray supervision scheme with a comprehensive ablation study. Finally, we show more results and applications of the proposed method. 

\noindent {\bf Dataset.} 
The aforementioned experiments are performed on three datasets.
First, we use the dataset released by DeepShadow\cite{DBLP:journals/corr/abs-2203-15065}, which contains binary shadow images of six scenes under different point lights.
Each scene is terrain-like and captured by a vertical-down camera.
For more complex scenes captured by other viewpoints, we find that no publicly available dataset satisfies our needs. 
Therefore, we construct new synthetic and real datasets for a thorough evaluation. 
For synthetic data, we render eight scenes using objects from the NeRF synthetic dataset\cite{DBLP:conf/eccv/MildenhallSTBRN20}.
Each test case is built by adding a horizontal plane to model the ground, placing the object on the plane, and rendering the scene using Blender\cite{blender}.
We render binary shadow images and RGB images of resolution 800$\times$800.
To test different light types, we render each scene with 100 directional lights and 100 point lights.
We select lights randomly sampled on the upper hemisphere, similar to the camera position selection in NeRF.
Our synthetic dataset features realistic materials with specular effects. 
Transparency and inter-reflections are disabled as these effects are beyond our assumption.
We also capture a real dataset to investigate our method's applicability to real capture setups.
For each scene, we place the object on the ground, illuminate the scene with only a handheld cellphone flashlight and capture it with a fixed camera.
We capture around 40 RGB images when the handheld flashlight moves around the scene and obtain the light locations similarly to \cite{DBLP:conf/eccv/BiXSHHKR20}.
We place a checkerboard on the ground and capture one additional image with the same fixed camera to calibrate the ground.
Please see \cref{tab:dataset_overview} for a summary of used datasets.

\begin{table}[!htp]\centering
  \scriptsize
  \begin{tabular}{lccccc}\toprule
  &RGB &Binary shadow &Directional light &Point light \\\midrule
  DeepShadow\cite{DBLP:journals/corr/abs-2203-15065} & &\checkmark & &\checkmark \\
  Our Synthetic &\checkmark &\checkmark &\checkmark &\checkmark \\
  Our Real &\checkmark & & &\checkmark \\
  \bottomrule
  \end{tabular}
  \caption{Datasets used in the evaluation.}\label{tab:dataset_overview}
  \vspace{-1em}
\end{table}

\noindent {\bf Metrics.}
As the compared methods output depth maps or normal maps of the visible regions, we also evaluate the quality of single-view reconstruction by depth errors in L1 (Depth L1) and normal errors in mean angular error (Normal MAE) computed in the visible foreground region.
It should be noted that as some compared methods output a depth map without a specific scale, Depth L1 is calculated
after aligning the depth map to the ground truth using ICP.

\begin{figure*}[t]
  \centering
  \begin{minipage}[t]{.48\textwidth}
    \centering
    \includeinkscape[width=.98\linewidth]{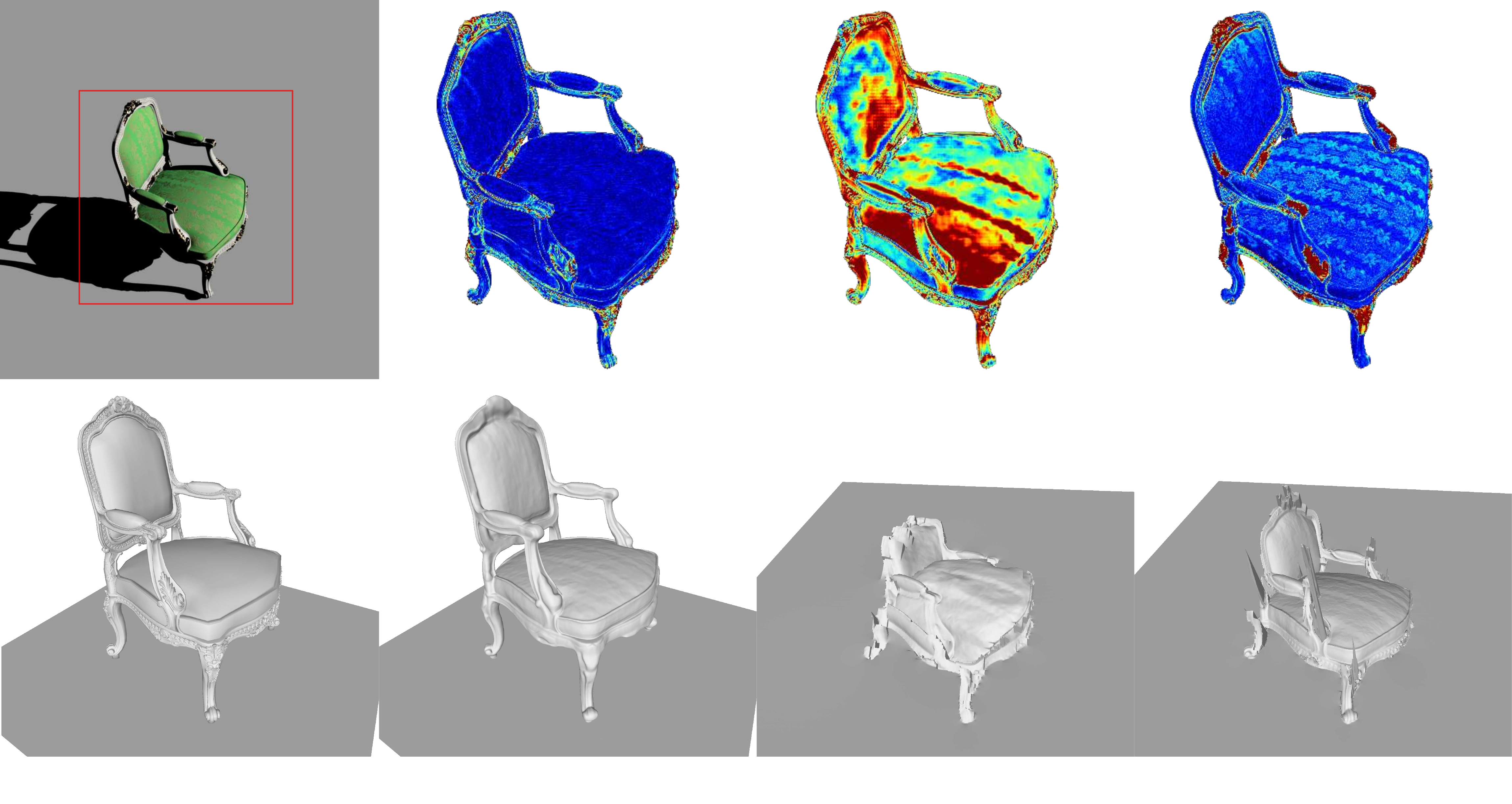}
  \end{minipage}
  \begin{minipage}[t]{.48\textwidth}
    \centering
    \includeinkscape[width=.98\linewidth]{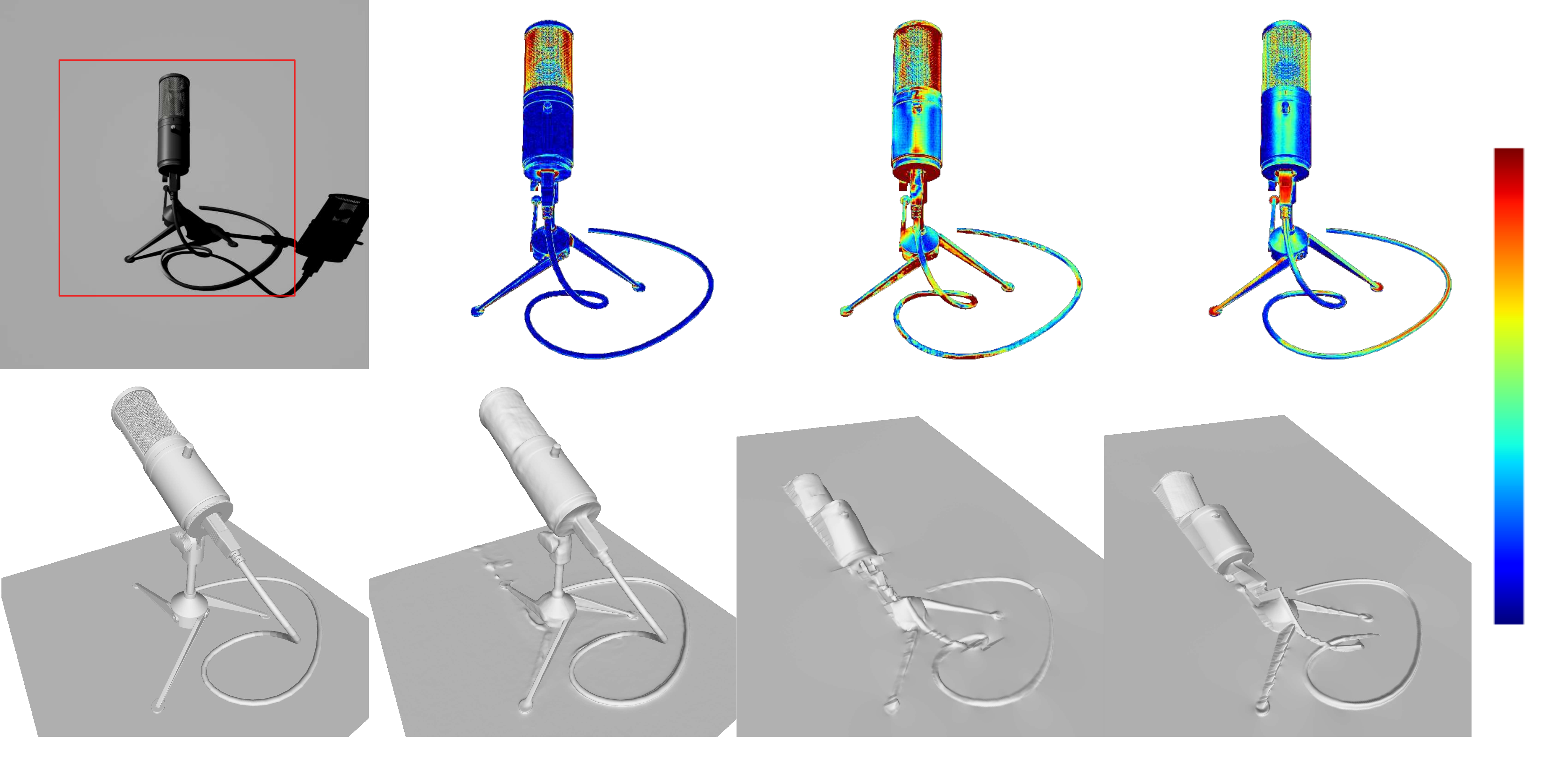}
  \end{minipage}
  \caption{Comparison on RGB inputs. The heat maps in the first row show the error distribution compared to the ground truth normal map.}
  \label{fig:rgb_quali}
  \vspace{-1em}
\end{figure*}

\begin{table}[t]\centering
  \footnotesize
  \begin{tabular}{lrrrrrrrrr}\toprule
  Method &Metric &DeepShadow dataset & Our binary \\\midrule
  DeepShadow &Depth L1$\downarrow$ &0.0223 &0.5020 \\
  Ours &Depth L1$\downarrow$ &\textbf{0.0135}  &\textbf{0.1870} \\\midrule
  DeepShadow &Normal MAE$\downarrow$ &20.93  &29.71\\
  Ours &Normal MAE$\downarrow$ &\textbf{19.68} &\textbf{20.21} \\
  \bottomrule
  \end{tabular}
  \caption{Quantitative comparison of reconstruction quality on the DeepShadow dataset and our binary shadow dataset.}\label{tab:shadow_quant_avg}
  \vspace{-1em}
  \end{table}

\subsubsection{Comparison on binary shadow inputs}
\label{sec:shadow_compare}

On binary shadow images, we compare our method with DeepShadow\cite{DBLP:journals/corr/abs-2203-15065}, the only existing method that supports scene reconstruction from similar inputs.
We find DeepShadow works better with a vertical-down camera, possibly because it represents the scene geometry as a depth map.
Therefore, we conduct this experiment on the DeepShadow dataset and the test samples captured under a similar viewpoint in our synthetic dataset. 
Although this setup gives advantages to DeepShadow, qualitative and quantitative results show that our method achieves better shape reconstruction on both datasets.
As shown at the top row of \cref{fig:shadow_quali}, our method achieves visually comparable results with DeepShadow on reconstructing a terrain-like geometry. 
For more complex inputs, our method reconstructs more detailed and complete structures than DeepShadow, as shown at the bottom row of \cref{fig:shadow_quali}. 
Benefiting from the shadow ray supervision of the complex shadow cast by the occluded geometry, our method can reconstruct the invisible regions, as shown by the results at the bottom right.
The results show that our method brings significant improvement in reconstructing complex scenes. 
Please see \cref{tab:shadow_quant_avg} for the quantitative results.
Note that our method requires the depth of the ground plane. 
This is also used by DeepShadow to initialize its depth map prediction.

\subsubsection{Comparison on RGB inputs}
\label{sec:rgb_compare}

On RGB inputs from our synthetic dataset, we compare our method with two state-of-the-art photometric stereo methods\cite{DBLP:conf/cvpr/ChenHSMW19,DBLP:conf/cvpr/LiL22} which also consider shadows.
SDPS-Net\cite{DBLP:conf/cvpr/ChenHSMW19} is a deep-learning method that augments the training dataset with images under shadows, and Li et al.\cite{DBLP:conf/cvpr/LiL22} is a recent neural field method that considers shadows cast by the reconstructed depth map. 
Both achieve higher performance in photometric stereo with the leverage of shadows.
Compared with these methods, our method can better leverage shape cues in the shadows to reconstruct shapes with more precise global structure as shown in \cref{fig:rgb_quali}.
Thanks to the shadow ray supervision of 3D neural SDF representation, our method can better handle abrupt depth changes at surface boundaries.
As shown in \cref{tab:ablation_study}, we achieve the lowest depth and normal reconstruction errors, illustrating the effectiveness of the proposed shadow ray supervision scheme in leveraging shadow information.

\begin{figure}[t]
  \centering
  \begin{minipage}[t]{.48\linewidth}
    \centering
    \includeinkscape[width=.98\linewidth]{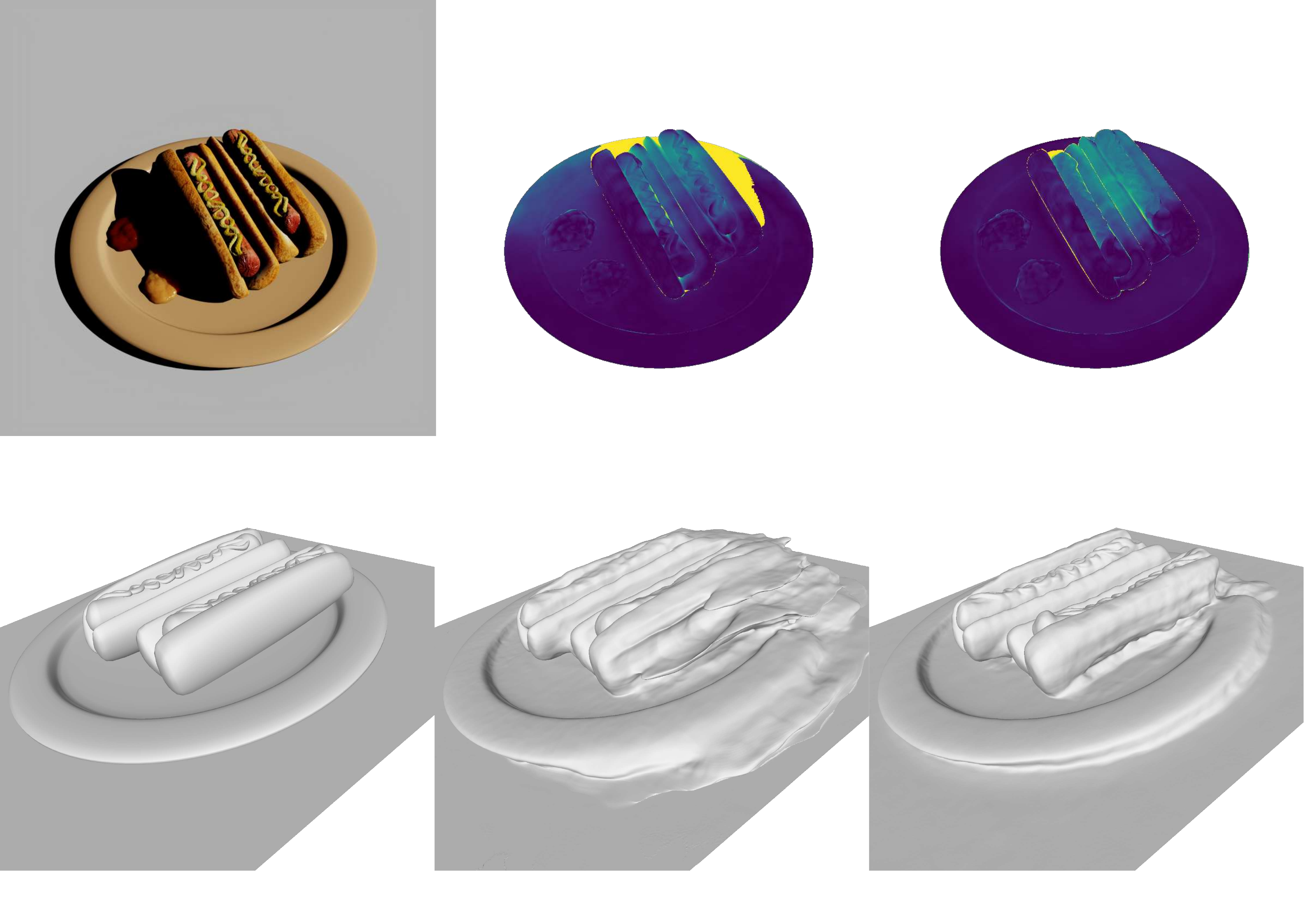}
  \end{minipage}
  \begin{minipage}[t]{.48\linewidth}
    \centering
    \includeinkscape[width=.98\linewidth]{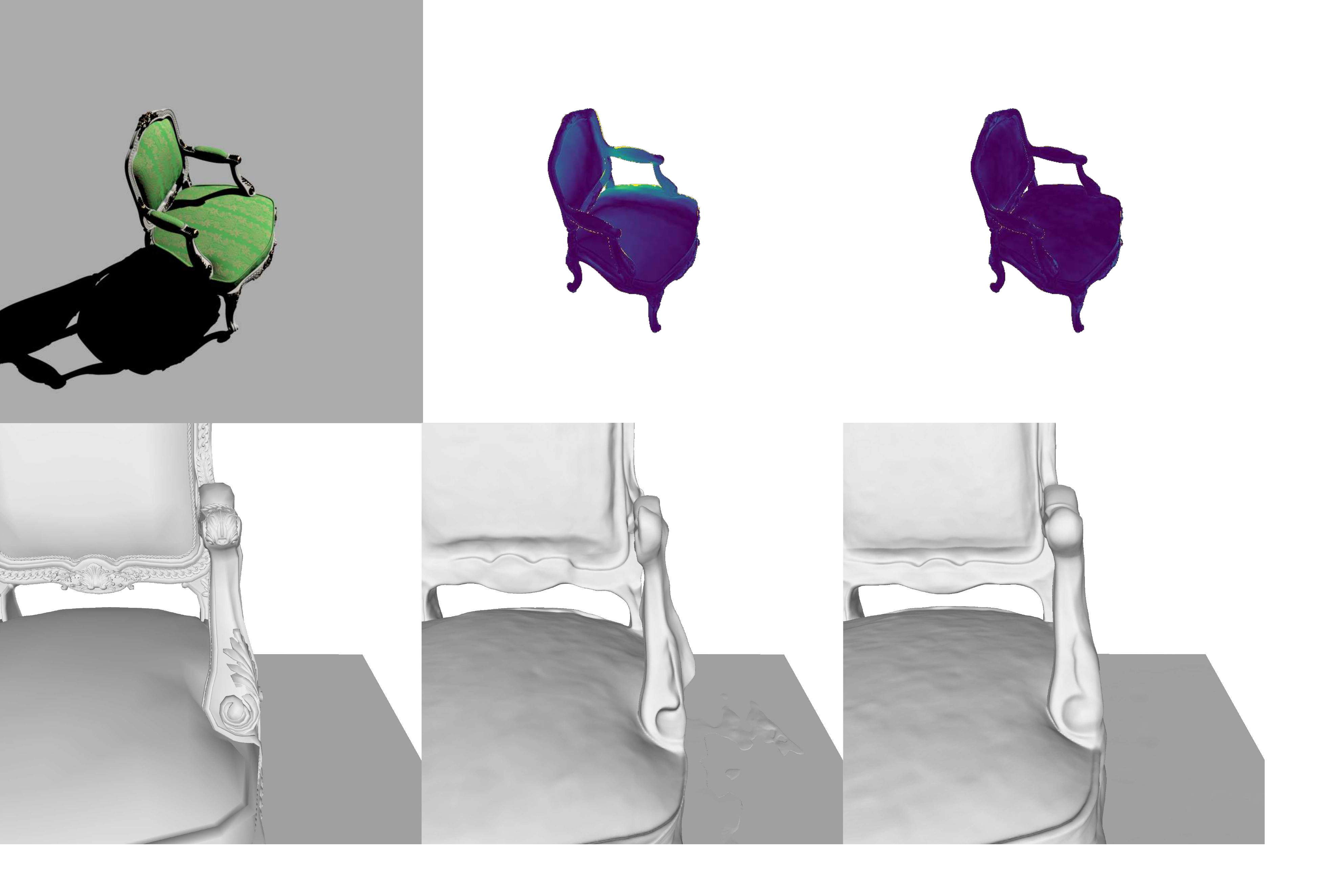}
  \end{minipage}
  \caption{Qualitative comparison with different ablations.}
  \label{fig:ablation_study}
\end{figure}

\begin{table}[t]\centering
  \footnotesize
  \begin{tabular}{lrrrrrrrrrrr}\toprule
  Method &Metric  &Avg &Metric  &Avg\\\midrule
  SDPS-Net &Depth L1$\downarrow$ &0.9163 &Normal MAE$\downarrow$ &38.94 \\
  Li et al. &Depth L1$\downarrow$ &0.8794  &Normal MAE$\downarrow$ &23.61\\
  W/o diff. inter. &Depth L1$\downarrow$  &0.2569  &Normal MAE$\downarrow$  &18.01\\
  W/o bound. samp. &Depth L1$\downarrow$  &0.3552 &Normal MAE$\downarrow$  &28.44\\
  Ours Full &Depth L1$\downarrow$  &\textbf{0.1341}  &Normal MAE$\downarrow$  &\textbf{15.03}\\
  \bottomrule
  \end{tabular}
  \caption{Quantitative results on our RGB dataset.}\label{tab:ablation_study}
  \vspace{-1em}
  \end{table}

\subsubsection{Ablation Study}

To demonstrate the effectiveness of the proposed differentiable intersection points and boundary sampling strategy, we construct two ablations by removing the two techniques and comparing them with our complete method on our synthetic directional RGB inputs.
In the first ablation, we only sample one shadow ray at a boundary pixel. 
As shown in the left half of \cref{fig:ablation_study}, without boundary sampling, the reconstructed geometry will extrude along the image plane direction, leading to significant errors around the boundary. 
In the second ablation, we directly use the non-differentiable intersection points. 
From the right half of  \cref{fig:ablation_study}, we can see that 
 the errors around the left arm increase as the network fail to update the depth using inaccurate backpropagated gradients. 
Quantitative results in \cref{tab:ablation_study} show that our proposed techniques greatly enhance the performance of geometry reconstruction.

\subsubsection{More Results}

\begin{figure}[t]
  \centering
  \includeinkscape[width=.98\linewidth]{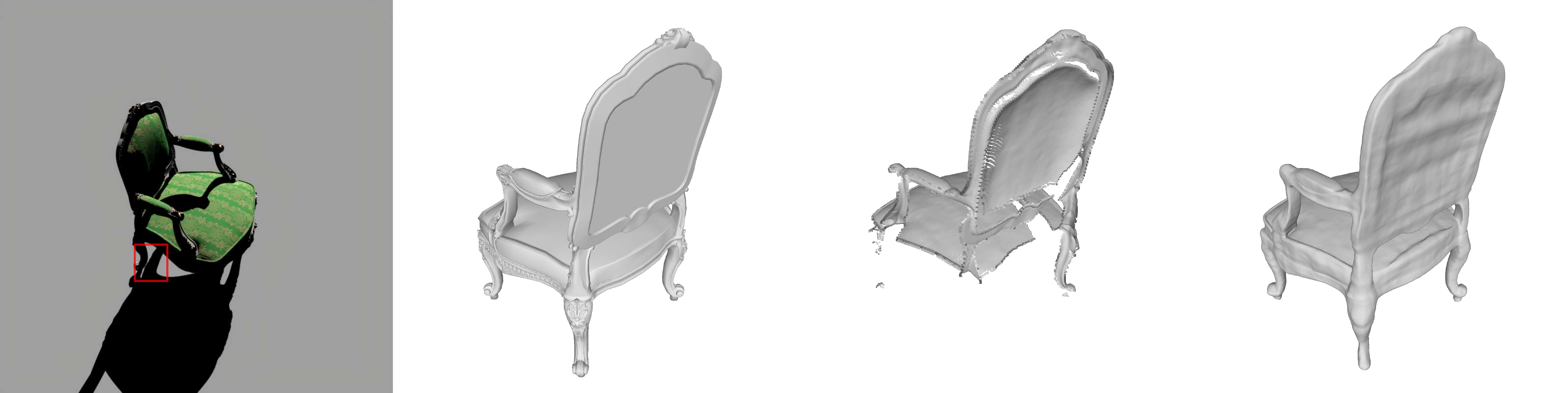}
  \includeinkscape[width=.98\linewidth]{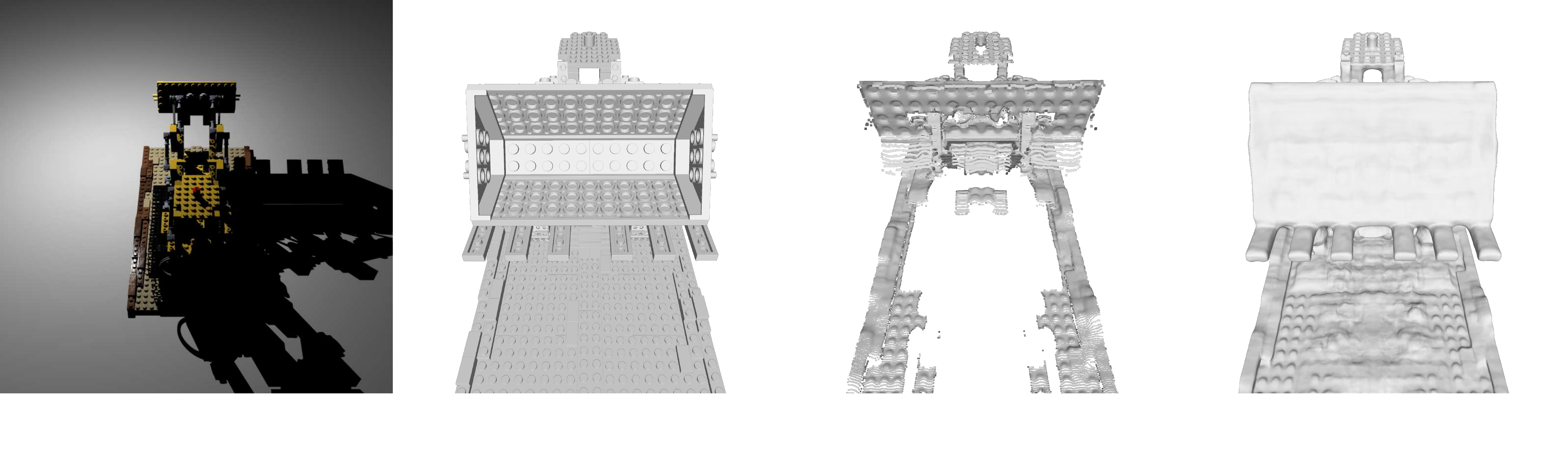}
  \caption{Results of invisible geometry reconstruction. The third column illustrates the region visible by the camera.}
  \label{fig:recon_invisble}
\end{figure}

\begin{figure}[t]
  \centering
  \begin{minipage}[t]{.48\linewidth}
    \centering
    \includeinkscape[width=.98\linewidth]{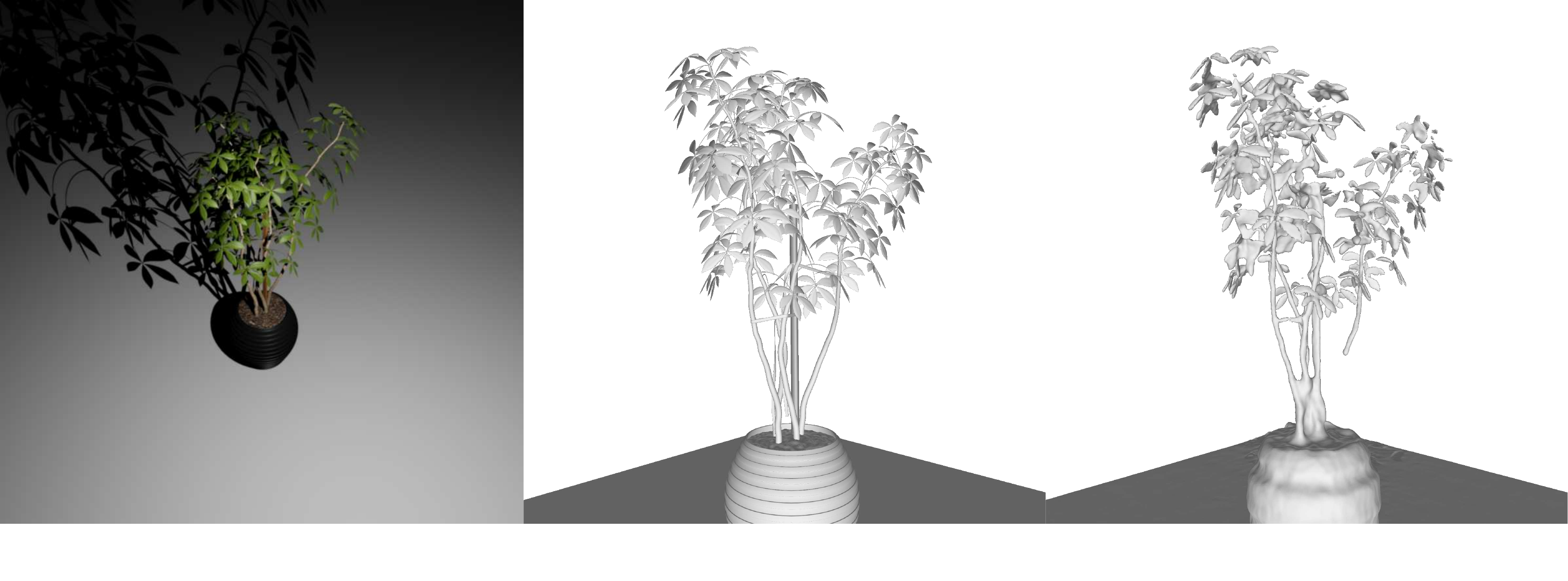}
  \end{minipage}
  \begin{minipage}[t]{.48\linewidth}
    \centering
    \includeinkscape[width=.98\linewidth]{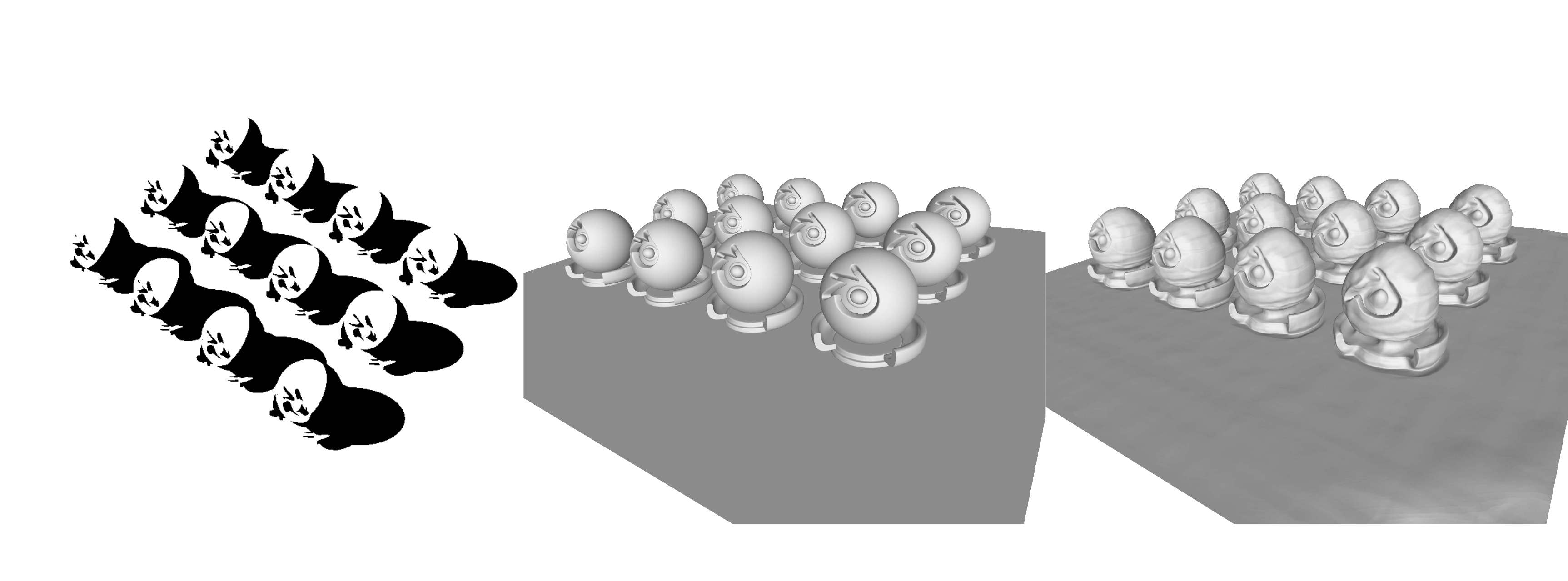}
  \end{minipage}
  \caption{Results on more various inputs.}
  \label{fig:other_inputs}
\end{figure}

We present more results to demonstrate the ability of the proposed method to reconstruct occluded geometry and synthesize images under novel lighting. We also test our method on handling more various inputs, including real images.

\noindent{\bf Reconstructing invisible geometry.}
Our method can reconstruct geometry that is not directly visible from the camera.
As shown in \cref{fig:recon_invisble}, our method reconstructs more complete geometry than the visible region in the third column, \eg, the invisible chair leg and the bulldozer blade.
As these invisible shapes cast shadows captured by the camera (labeled by red boxes in \cref{fig:recon_invisble}), the corresponding shadow rays can supervise the shape to match the shadows.

\begin{figure}[t]
  \centering
  \includeinkscape[width=.98\linewidth]{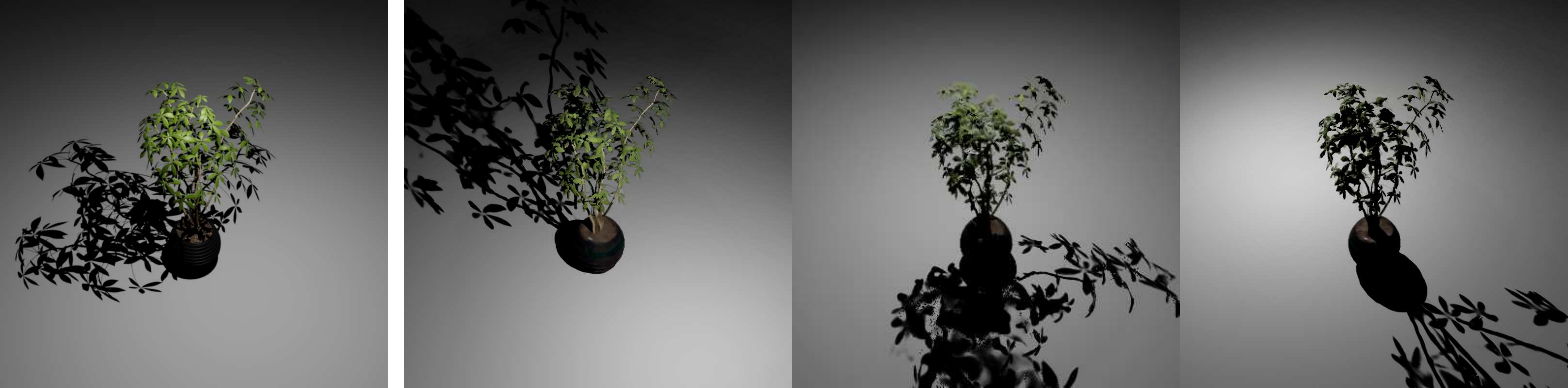}
  \includeinkscape[width=.98\linewidth]{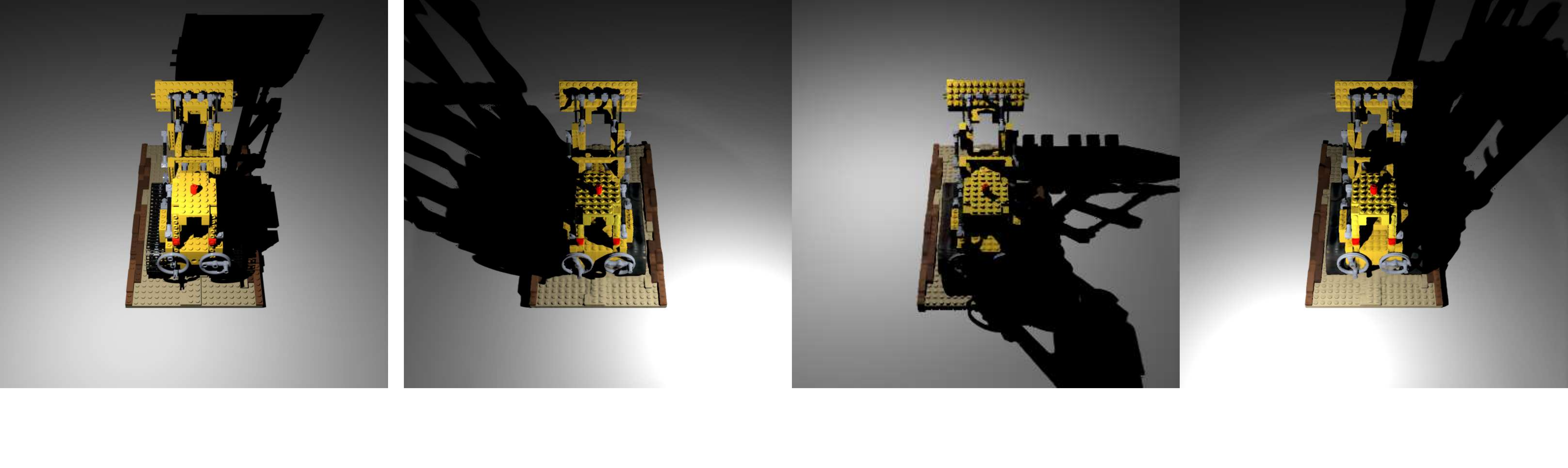}
  \caption{Results of novel-light synthesis.}
  \label{fig:relight}
\end{figure}

\noindent {\bf Novel-light synthesis.}
After reconstructing the neural scene, we can re-render the scene under a novel light direction, as shown in \cref{fig:relight}.
Besides shading and specular effects, we can generate accurate shadows on the ground and the object itself, consistent with the object's shape. 
The results also indicate that it is beneficial to integrate shadow ray supervision into a neural relighting pipeline.
Please also see the supplementary video for continuous relighting results.

\begin{figure}[t]
  \centering
  \includeinkscape[width=.98\linewidth]{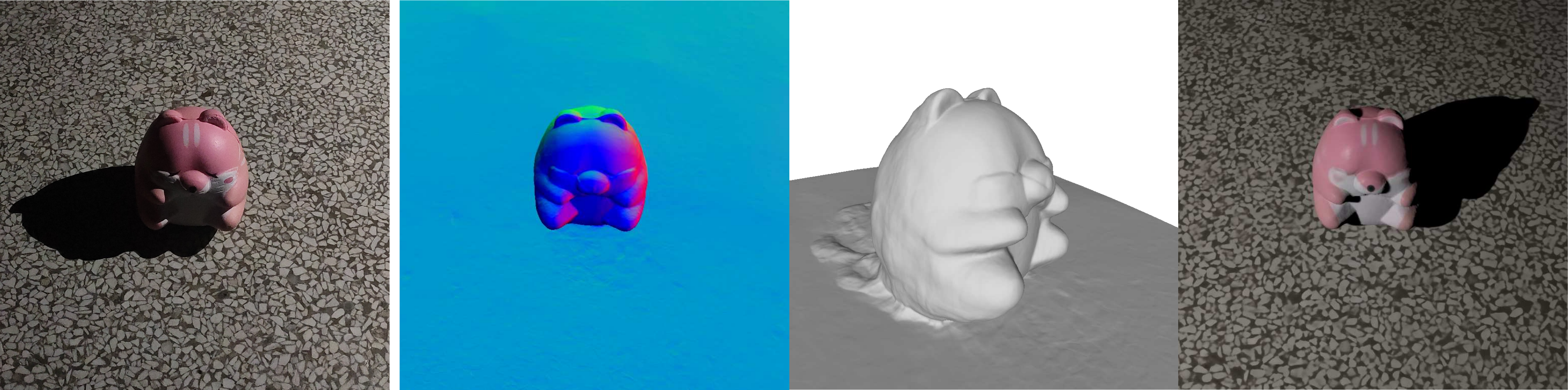}
  \includeinkscape[width=.98\linewidth]{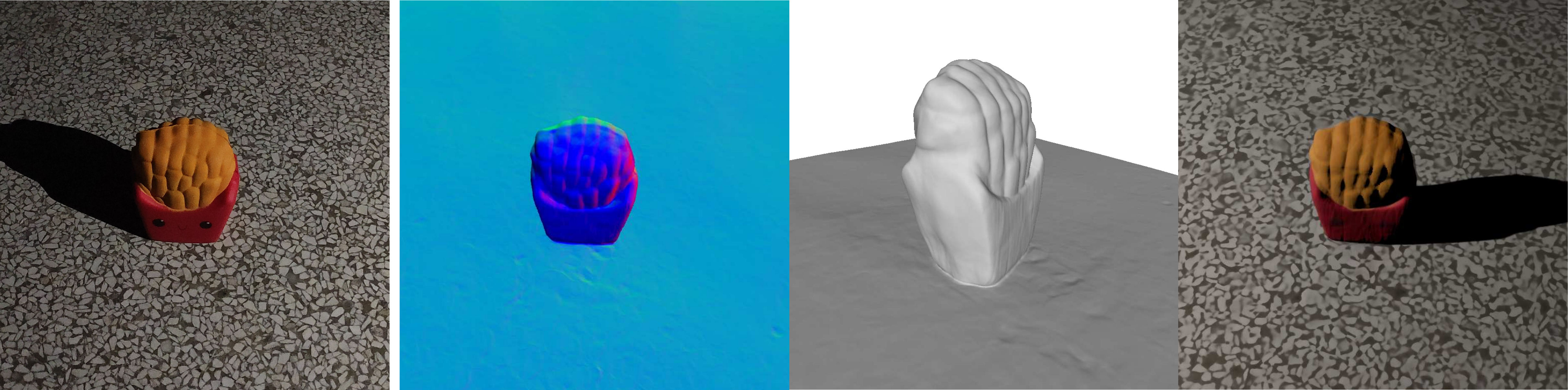}
  \includeinkscape[width=.98\linewidth]{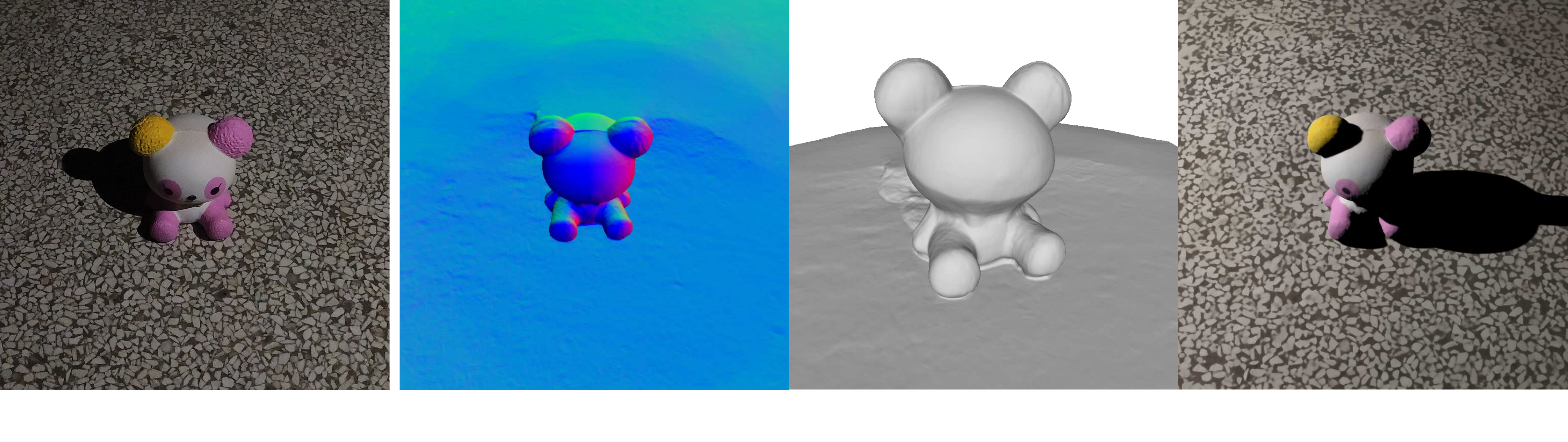}
  \caption{Results on real data.}
  \label{fig:real_results}
\end{figure}

\noindent {\bf Results on more various inputs.}
In order to demonstrate the generalization of our method, we test our method on more challenging synthetic data.
As shown in \cref{fig:other_inputs}, our method can reconstruct scenes with multiple objects (Column 2). Our method still successfully reconstructs some leaves and stems for inputs with extremely complex structures for single-view reconstruction (Column 1). We further apply our method to our real data. 
As shown in \cref{fig:real_results}, our method reconstructs complete 3D shapes and accurate surface details from the simple setup and can handle the ground with non-trivial materials. Reconstructed results from real inputs can also generate realistic relighting results.

\subsection{Limitations}%

The effectiveness of the proposed shadow ray supervision in reconstructing neural scenes is demonstrated by extensive experiments. 
However, as an early attempt to model shadow rays, our method is based on several assumptions.
We assume the scene does not emit light and ignore inter-reflections to simplify light modeling.
We observe that some thin structures are too complex that they can still be missing in our reconstruction.
It is a general limitation and can be improved by the progress in thin structure neural SDF, as indicated by very recent works\cite{DBLP:journals/corr/abs-2211-14173,DBLP:conf/cvpr/0001LLY22}.

\section{Conclusion}
Compared with NeRF supervising camera rays, we achieve fully differentiable supervision of shadow rays in a neural scene representation.
This technique enables shape reconstruction from single-view multi-light observations and supports both pure shadow and RGB inputs.
Our technique works well for both point and directional lights and can be used for 3D reconstruction and relighting.
A multi-ray sampling strategy is proposed to handle challenges posed by surface boundaries in locating shadow rays.
Experiments show that our technique outperforms the SOTAs in single-view reconstruction, and it has the power to reconstruct scene geometries out of the camera's line of sight.

\setlength\parindent{0pt}
{\bf Acknowledgements}
This work was supported by the National Key R\&D Program of China (2018YFA0704000), Beijing Natural Science Foundation (M22024), the NSFC (No.62021002), and the Key Research and Development Project of Tibet Autonomous Region (XZ202101ZY0019G). This work was also supported by THUIBCS, Tsinghua University, and BLBCI, Beijing Municipal Education Commission.

{\small
\bibliographystyle{ieee_fullname}
\bibliography{egbib}

\begin{thebibliography}{10}\itemsep=-1pt

\bibitem{DBLP:conf/nips/AtzmonHYIML19}
Matan Atzmon, Niv Haim, Lior Yariv, Ofer Israelov, Haggai Maron, and Yaron
  Lipman.
\newblock Controlling neural level sets.
\newblock In Hanna~M. Wallach, Hugo Larochelle, Alina Beygelzimer, Florence
  d'Alch{\'{e}}{-}Buc, Emily~B. Fox, and Roman Garnett, editors, {\em Advances
  in Neural Information Processing Systems 32: Annual Conference on Neural
  Information Processing Systems 2019, NeurIPS 2019, December 8-14, 2019,
  Vancouver, BC, Canada}, pages 2032--2041, 2019.

\bibitem{DBLP:conf/siggrapha/BangaruGLLSHBXB22}
Sai~Praveen Bangaru, Micha{\"{e}}l Gharbi, Fujun Luan, Tzu{-}Mao Li, Kalyan
  Sunkavalli, Milos Hasan, Sai Bi, Zexiang Xu, Gilbert Bernstein, and
  Fr{\'{e}}do Durand.
\newblock Differentiable rendering of neural sdfs through reparameterization.
\newblock In Soon~Ki Jung, Jehee Lee, and Adam~W. Bargteil, editors, {\em
  {SIGGRAPH} Asia 2022 Conference Papers, {SA} 2022, Daegu, Republic of Korea,
  December 6-9, 2022}, pages 5:1--5:9. {ACM}, 2022.

\bibitem{DBLP:journals/corr/abs-2008-03824}
Sai Bi, Zexiang Xu, Pratul~P. Srinivasan, Ben Mildenhall, Kalyan Sunkavalli,
  Milos Hasan, Yannick Hold{-}Geoffroy, David~J. Kriegman, and Ravi
  Ramamoorthi.
\newblock Neural reflectance fields for appearance acquisition.
\newblock {\em CoRR}, abs/2008.03824, 2020.

\bibitem{DBLP:conf/eccv/BiXSHHKR20}
Sai Bi, Zexiang Xu, Kalyan Sunkavalli, Milos Hasan, Yannick Hold{-}Geoffroy,
  David~J. Kriegman, and Ravi Ramamoorthi.
\newblock Deep reflectance volumes: Relightable reconstructions from multi-view
  photometric images.
\newblock In Andrea Vedaldi, Horst Bischof, Thomas Brox, and Jan{-}Michael
  Frahm, editors, {\em Computer Vision - {ECCV} 2020 - 16th European
  Conference, Glasgow, UK, August 23-28, 2020, Proceedings, Part {III}}, volume
  12348 of {\em Lecture Notes in Computer Science}, pages 294--311. Springer,
  2020.

\bibitem{DBLP:conf/iccv/BossBJBLL21}
Mark Boss, Raphael Braun, Varun Jampani, Jonathan~T. Barron, Ce Liu, and
  Hendrik P.~A. Lensch.
\newblock Nerd: Neural reflectance decomposition from image collections.
\newblock In {\em 2021 {IEEE/CVF} International Conference on Computer Vision,
  {ICCV} 2021, Montreal, QC, Canada, October 10-17, 2021}, pages 12664--12674.
  {IEEE}, 2021.

\bibitem{DBLP:conf/nips/BossJBLBL21}
Mark Boss, Varun Jampani, Raphael Braun, Ce Liu, Jonathan~T. Barron, and
  Hendrik P.~A. Lensch.
\newblock Neural-pil: Neural pre-integrated lighting for reflectance
  decomposition.
\newblock In Marc'Aurelio Ranzato, Alina Beygelzimer, Yann~N. Dauphin, Percy
  Liang, and Jennifer~Wortman Vaughan, editors, {\em Advances in Neural
  Information Processing Systems 34: Annual Conference on Neural Information
  Processing Systems 2021, NeurIPS 2021, December 6-14, 2021, virtual}, pages
  10691--10704, 2021.

\bibitem{DBLP:conf/eccv/CaoSSOM22}
Xu Cao, Hiroaki Santo, Boxin Shi, Fumio Okura, and Yasuyuki Matsushita.
\newblock Bilateral normal integration.
\newblock In Shai Avidan, Gabriel~J. Brostow, Moustapha Ciss{\'{e}},
  Giovanni~Maria Farinella, and Tal Hassner, editors, {\em Computer Vision -
  {ECCV} 2022 - 17th European Conference, Tel Aviv, Israel, October 23-27,
  2022, Proceedings, Part {I}}, volume 13661 of {\em Lecture Notes in Computer
  Science}, pages 552--567. Springer, 2022.

\bibitem{DBLP:conf/cvpr/CaoSOM21}
Xu Cao, Boxin Shi, Fumio Okura, and Yasuyuki Matsushita.
\newblock Normal integration via inverse plane fitting with minimum
  point-to-plane distance.
\newblock In {\em {IEEE} Conference on Computer Vision and Pattern Recognition,
  {CVPR} 2021, virtual, June 19-25, 2021}, pages 2382--2391. Computer Vision
  Foundation / {IEEE}, 2021.

\bibitem{DBLP:conf/cvpr/ChenHSMW19}
Guanying Chen, Kai Han, Boxin Shi, Yasuyuki Matsushita, and Kwan{-}Yee~K. Wong.
\newblock Self-calibrating deep photometric stereo networks.
\newblock In {\em {IEEE} Conference on Computer Vision and Pattern Recognition,
  {CVPR} 2019, Long Beach, CA, USA, June 16-20, 2019}, pages 8739--8747.
  Computer Vision Foundation / {IEEE}, 2019.

\bibitem{DBLP:conf/cvpr/0001LLY22}
Weikai Chen, Cheng Lin, Weiyang Li, and Bo Yang.
\newblock 3psdf: Three-pole signed distance function for learning surfaces with
  arbitrary topologies.
\newblock In {\em {IEEE/CVF} Conference on Computer Vision and Pattern
  Recognition, {CVPR} 2022, New Orleans, LA, USA, June 18-24, 2022}, pages
  18501--18510. {IEEE}, 2022.

\bibitem{DBLP:conf/eccv/ChenL22}
Zhaoxi Chen and Ziwei Liu.
\newblock Relighting4d: Neural relightable human from videos.
\newblock In Shai Avidan, Gabriel~J. Brostow, Moustapha Ciss{\'{e}},
  Giovanni~Maria Farinella, and Tal Hassner, editors, {\em Computer Vision -
  {ECCV} 2022 - 17th European Conference, Tel Aviv, Israel, October 23-27,
  2022, Proceedings, Part {XIV}}, volume 13674 of {\em Lecture Notes in
  Computer Science}, pages 606--623. Springer, 2022.

\bibitem{DBLP:journals/corr/abs-2204-05735}
Shin{-}Fang Chng, Sameera Ramasinghe, Jamie Sherrah, and Simon Lucey.
\newblock {GARF:} gaussian activated radiance fields for high fidelity
  reconstruction and pose estimation.
\newblock {\em CoRR}, abs/2204.05735, 2022.

\bibitem{blender}
Blender~Online Community.
\newblock {\em Blender - a 3D modelling and rendering package}.
\newblock Blender Foundation, Stichting Blender Foundation, Amsterdam, 2018.

\bibitem{DBLP:conf/cvpr/DaumD98}
Michael Daum and Gregory Dudek.
\newblock On 3-d surface reconstruction using shape from shadows.
\newblock In {\em 1998 Conference on Computer Vision and Pattern Recognition
  {(CVPR} '98), June 23-25, 1998, Santa Barbara, CA, {USA}}, pages 461--468.
  {IEEE} Computer Society, 1998.

\bibitem{DBLP:conf/icml/GroppYHAL20}
Amos Gropp, Lior Yariv, Niv Haim, Matan Atzmon, and Yaron Lipman.
\newblock Implicit geometric regularization for learning shapes.
\newblock In {\em Proceedings of the 37th International Conference on Machine
  Learning, {ICML} 2020, 13-18 July 2020, Virtual Event}, volume 119 of {\em
  Proceedings of Machine Learning Research}, pages 3789--3799. {PMLR}, 2020.

\bibitem{DBLP:conf/cvpr/HatzitheodorouK88}
Michael Hatzitheodorou and John~R. Kender.
\newblock An optimal algorithm for the derivation of shape from shadows.
\newblock In {\em {IEEE} Computer Society Conference on Computer Vision and
  Pattern Recognition, {CVPR} 1988, 5-9 June, 1988, Ann Arbor, Michigan,
  {USA}}, pages 486--491. {IEEE}, 1988.

\bibitem{DBLP:conf/iccv/JainTA21}
Ajay Jain, Matthew Tancik, and Pieter Abbeel.
\newblock Putting nerf on a diet: Semantically consistent few-shot view
  synthesis.
\newblock In {\em 2021 {IEEE/CVF} International Conference on Computer Vision,
  {ICCV} 2021, Montreal, QC, Canada, October 10-17, 2021}, pages 5865--5874.
  {IEEE}, 2021.

\bibitem{DBLP:journals/corr/abs-2203-15065}
Asaf Karnieli, Ohad Fried, and Yacov Hel{-}Or.
\newblock Deepshadow: Neural shape from shadow.
\newblock {\em CoRR}, abs/2203.15065, 2022.

\bibitem{DBLP:conf/aaai/KenderS86}
John~R. Kender and Earl Smith.
\newblock Shape from darkness: Deriving surface information from dynamic
  shadows.
\newblock In Tom Kehler, editor, {\em Proceedings of the 5th National
  Conference on Artificial Intelligence. Philadelphia, PA, USA, August 11-15,
  1986. Volume 1: Science}, pages 664--669. Morgan Kaufmann, 1986.

\bibitem{DBLP:journals/corr/KingmaB14}
Diederik~P. Kingma and Jimmy Ba.
\newblock Adam: {A} method for stochastic optimization.
\newblock In Yoshua Bengio and Yann LeCun, editors, {\em 3rd International
  Conference on Learning Representations, {ICLR} 2015, San Diego, CA, USA, May
  7-9, 2015, Conference Track Proceedings}, 2015.

\bibitem{DBLP:journals/tog/LaineHKSLA20}
Samuli Laine, Janne Hellsten, Tero Karras, Yeongho Seol, Jaakko Lehtinen, and
  Timo Aila.
\newblock Modular primitives for high-performance differentiable rendering.
\newblock {\em {ACM} Trans. Graph.}, 39(6):194:1--194:14, 2020.

\bibitem{DBLP:conf/iros/LangerDZ95}
Michael~S. Langer, Gregory Dudek, and Steven~W. Zucker.
\newblock Space occupancy using multiple shadowimages.
\newblock In {\em Proceedings of {IEEE/RSJ} International Conference on
  Intelligent Robots and Systems, {IROS} 1995, August 5 - 9, 1995, Pittsburgh,
  PA, {USA}}, pages 285--290. {IEEE} Computer Society, 1995.

\bibitem{DBLP:conf/cvpr/LiL22}
Junxuan Li and Hongdong Li.
\newblock Neural reflectance for shape recovery with shadow handling.
\newblock In {\em {IEEE/CVF} Conference on Computer Vision and Pattern
  Recognition, {CVPR} 2022, New Orleans, LA, USA, June 18-24, 2022}, pages
  16200--16209. {IEEE}, 2022.

\bibitem{DBLP:journals/tog/LiADL18}
Tzu{-}Mao Li, Miika Aittala, Fr{\'{e}}do Durand, and Jaakko Lehtinen.
\newblock Differentiable monte carlo ray tracing through edge sampling.
\newblock {\em {ACM} Trans. Graph.}, 37(6):222, 2018.

\bibitem{DBLP:conf/iccv/LinM0L21}
Chen{-}Hsuan Lin, Wei{-}Chiu Ma, Antonio Torralba, and Simon Lucey.
\newblock {BARF:} bundle-adjusting neural radiance fields.
\newblock In {\em 2021 {IEEE/CVF} International Conference on Computer Vision,
  {ICCV} 2021, Montreal, QC, Canada, October 10-17, 2021}, pages 5721--5731.
  {IEEE}, 2021.

\bibitem{10.1145/3306346.3323020}
Stephen Lombardi, Tomas Simon, Jason Saragih, Gabriel Schwartz, Andreas
  Lehrmann, and Yaser Sheikh.
\newblock Neural volumes: Learning dynamic renderable volumes from images.
\newblock {\em ACM Trans. Graph.}, 38(4), jul 2019.

\bibitem{DBLP:journals/corr/abs-2211-14173}
Xiaoxiao Long, Cheng Lin, Lingjie Liu, Yuan Liu, Peng Wang, Christian Theobalt,
  Taku Komura, and Wenping Wang.
\newblock Neuraludf: Learning unsigned distance fields for multi-view
  reconstruction of surfaces with arbitrary topologies.
\newblock {\em CoRR}, abs/2211.14173, 2022.

\bibitem{DBLP:conf/cvpr/MeschederONNG19}
Lars~M. Mescheder, Michael Oechsle, Michael Niemeyer, Sebastian Nowozin, and
  Andreas Geiger.
\newblock Occupancy networks: Learning 3d reconstruction in function space.
\newblock In {\em {IEEE} Conference on Computer Vision and Pattern Recognition,
  {CVPR} 2019, Long Beach, CA, USA, June 16-20, 2019}, pages 4460--4470.
  Computer Vision Foundation / {IEEE}, 2019.

\bibitem{DBLP:conf/eccv/MildenhallSTBRN20}
Ben Mildenhall, Pratul~P. Srinivasan, Matthew Tancik, Jonathan~T. Barron, Ravi
  Ramamoorthi, and Ren Ng.
\newblock Nerf: Representing scenes as neural radiance fields for view
  synthesis.
\newblock In Andrea Vedaldi, Horst Bischof, Thomas Brox, and Jan{-}Michael
  Frahm, editors, {\em Computer Vision - {ECCV} 2020 - 16th European
  Conference, Glasgow, UK, August 23-28, 2020, Proceedings, Part {I}}, volume
  12346 of {\em Lecture Notes in Computer Science}, pages 405--421. Springer,
  2020.

\bibitem{DBLP:conf/cvpr/NiemeyerMOG20}
Michael Niemeyer, Lars~M. Mescheder, Michael Oechsle, and Andreas Geiger.
\newblock Differentiable volumetric rendering: Learning implicit 3d
  representations without 3d supervision.
\newblock In {\em 2020 {IEEE/CVF} Conference on Computer Vision and Pattern
  Recognition, {CVPR} 2020, Seattle, WA, USA, June 13-19, 2020}, pages
  3501--3512. Computer Vision Foundation / {IEEE}, 2020.

\bibitem{DBLP:conf/iccv/OechsleP021}
Michael Oechsle, Songyou Peng, and Andreas Geiger.
\newblock {UNISURF:} unifying neural implicit surfaces and radiance fields for
  multi-view reconstruction.
\newblock In {\em 2021 {IEEE/CVF} International Conference on Computer Vision,
  {ICCV} 2021, Montreal, QC, Canada, October 10-17, 2021}, pages 5569--5579.
  {IEEE}, 2021.

\bibitem{DBLP:journals/nature/OTooleLW18}
Matthew O'Toole, David~B. Lindell, and Gordon Wetzstein.
\newblock Confocal non-line-of-sight imaging based on the light-cone transform.
\newblock {\em Nat.}, 555(7696):338--341, 2018.

\bibitem{DBLP:conf/cvpr/ParkFSNL19}
Jeong~Joon Park, Peter Florence, Julian Straub, Richard~A. Newcombe, and Steven
  Lovegrove.
\newblock Deepsdf: Learning continuous signed distance functions for shape
  representation.
\newblock In {\em {IEEE} Conference on Computer Vision and Pattern Recognition,
  {CVPR} 2019, Long Beach, CA, USA, June 16-20, 2019}, pages 165--174. Computer
  Vision Foundation / {IEEE}, 2019.

\bibitem{DBLP:journals/trob/RavivPL89}
Daniel Raviv, Yoh{-}Han Pao, and Kenneth~A. Loparo.
\newblock Reconstruction of three-dimensional surfaces from two-dimensional
  binary images.
\newblock {\em {IEEE} Trans. Robotics Autom.}, 5(5):701--710, 1989.

\bibitem{DBLP:journals/corr/abs-2112-05140}
Viktor Rudnev, Mohamed Elgharib, William A.~P. Smith, Lingjie Liu, Vladislav
  Golyanik, and Christian Theobalt.
\newblock Neural radiance fields for outdoor scene relighting.
\newblock {\em CoRR}, abs/2112.05140, 2021.

\bibitem{DBLP:journals/ijcv/SavareseARBP07}
Silvio Savarese, Marco Andreetto, Holly~E. Rushmeier, Fausto Bernardini, and
  Pietro Perona.
\newblock 3d reconstruction by shadow carving: Theory and practical evaluation.
\newblock {\em Int. J. Comput. Vis.}, 71(3):305--336, 2007.

\bibitem{DBLP:conf/iccv/SavareseRBP01}
Silvio Savarese, Holly~E. Rushmeier, Fausto Bernardini, and Pietro Perona.
\newblock Shadow carving.
\newblock In {\em Proceedings of the Eighth International Conference On
  Computer Vision (ICCV-01), Vancouver, British Columbia, Canada, July 7-14,
  2001 - Volume 1}, pages 190--197. {IEEE} Computer Society, 2001.

\bibitem{DBLP:journals/pami/ShenWLPLGLY21}
Siyuan Shen, Zi Wang, Ping Liu, Zhengqing Pan, Ruiqian Li, Tian Gao, Shiying
  Li, and Jingyi Yu.
\newblock Non-line-of-sight imaging via neural transient fields.
\newblock {\em {IEEE} Trans. Pattern Anal. Mach. Intell.}, 43(7):2257--2268,
  2021.

\bibitem{DBLP:conf/cvpr/SrinivasanDZTMB21}
Pratul~P. Srinivasan, Boyang Deng, Xiuming Zhang, Matthew Tancik, Ben
  Mildenhall, and Jonathan~T. Barron.
\newblock Nerv: Neural reflectance and visibility fields for relighting and
  view synthesis.
\newblock In {\em {IEEE} Conference on Computer Vision and Pattern Recognition,
  {CVPR} 2021, virtual, June 19-25, 2021}, pages 7495--7504. Computer Vision
  Foundation / {IEEE}, 2021.

\bibitem{DBLP:journals/tog/SteinbergY21}
Shlomi Steinberg and Ling{-}Qi Yan.
\newblock A generic framework for physical light transport.
\newblock {\em {ACM} Trans. Graph.}, 40(4):139:1--139:20, 2021.

\bibitem{DBLP:journals/corr/abs-2203-15946}
Kushagra Tiwary, Tzofi Klinghoffer, and Ramesh Raskar.
\newblock Towards learning neural representations from shadows.
\newblock {\em CoRR}, abs/2203.15946, 2022.

\bibitem{DBLP:journals/tog/ViciniSJ22}
Delio Vicini, S{\'{e}}bastien Speierer, and Wenzel Jakob.
\newblock Differentiable signed distance function rendering.
\newblock {\em {ACM} Trans. Graph.}, 41(4):125:1--125:18, 2022.

\bibitem{DBLP:conf/nips/WangLLTKW21}
Peng Wang, Lingjie Liu, Yuan Liu, Christian Theobalt, Taku Komura, and Wenping
  Wang.
\newblock Neus: Learning neural implicit surfaces by volume rendering for
  multi-view reconstruction.
\newblock In Marc'Aurelio Ranzato, Alina Beygelzimer, Yann~N. Dauphin, Percy
  Liang, and Jennifer~Wortman Vaughan, editors, {\em Advances in Neural
  Information Processing Systems 34: Annual Conference on Neural Information
  Processing Systems 2021, NeurIPS 2021, December 6-14, 2021, virtual}, pages
  27171--27183, 2021.

\bibitem{DBLP:journals/corr/abs-2102-07064}
Zirui Wang, Shangzhe Wu, Weidi Xie, Min Chen, and Victor~Adrian Prisacariu.
\newblock Nerf-: Neural radiance fields without known camera parameters.
\newblock {\em CoRR}, abs/2102.07064, 2021.

\bibitem{DBLP:journals/cgf/XieTSLYKTTSS22}
Yiheng Xie, Towaki Takikawa, Shunsuke Saito, Or Litany, Shiqin Yan, Numair
  Khan, Federico Tombari, James Tompkin, Vincent Sitzmann, and Srinath Sridhar.
\newblock Neural fields in visual computing and beyond.
\newblock {\em Comput. Graph. Forum}, 41(2):641--676, 2022.

\bibitem{DBLP:conf/cvpr/XinNKSNG19}
Shumian Xin, Sotiris Nousias, Kiriakos~N. Kutulakos, Aswin~C. Sankaranarayanan,
  Srinivasa~G. Narasimhan, and Ioannis Gkioulekas.
\newblock A theory of fermat paths for non-line-of-sight shape reconstruction.
\newblock In {\em {IEEE} Conference on Computer Vision and Pattern Recognition,
  {CVPR} 2019, Long Beach, CA, USA, June 16-20, 2019}, pages 6800--6809.
  Computer Vision Foundation / {IEEE}, 2019.

\bibitem{DBLP:conf/eccv/XuJWFSW22}
Dejia Xu, Yifan Jiang, Peihao Wang, Zhiwen Fan, Humphrey Shi, and Zhangyang
  Wang.
\newblock Sinnerf: Training neural radiance fields on complex scenes from a
  single image.
\newblock In Shai Avidan, Gabriel~J. Brostow, Moustapha Ciss{\'{e}},
  Giovanni~Maria Farinella, and Tal Hassner, editors, {\em Computer Vision -
  {ECCV} 2022 - 17th European Conference, Tel Aviv, Israel, October 23-27,
  2022, Proceedings, Part {XXII}}, volume 13682 of {\em Lecture Notes in
  Computer Science}, pages 736--753. Springer, 2022.

\bibitem{DBLP:journals/ijcv/YamazakiNBK09}
Shuntaro Yamazaki, Srinivasa~G. Narasimhan, Simon Baker, and Takeo Kanade.
\newblock The theory and practice of coplanar shadowgram imaging for acquiring
  visual hulls of intricate objects.
\newblock {\em Int. J. Comput. Vis.}, 81(3):259--280, 2009.

\bibitem{DBLP:conf/eccv/YangCCCW22}
Wenqi Yang, Guanying Chen, Chaofeng Chen, Zhenfang Chen, and Kwan{-}Yee~K.
  Wong.
\newblock Ps-nerf: Neural inverse rendering for multi-view photometric stereo.
\newblock In Shai Avidan, Gabriel~J. Brostow, Moustapha Ciss{\'{e}},
  Giovanni~Maria Farinella, and Tal Hassner, editors, {\em Computer Vision -
  {ECCV} 2022 - 17th European Conference, Tel Aviv, Israel, October 23-27,
  2022, Proceedings, Part {I}}, volume 13661 of {\em Lecture Notes in Computer
  Science}, pages 266--284. Springer, 2022.

\bibitem{DBLP:journals/corr/abs-2210-08936}
Wenqi Yang, Guanying Chen, Chaofeng Chen, Zhenfang Chen, and Kwan{-}Yee~K.
  Wong.
\newblock S\({}^{\mbox{3}}\)-nerf: Neural reflectance field from shading and
  shadow under a single viewpoint.
\newblock {\em CoRR}, abs/2210.08936, 2022.

\bibitem{DBLP:conf/eccv/YaoZLQFMTQ22}
Yao Yao, Jingyang Zhang, Jingbo Liu, Yihang Qu, Tian Fang, David McKinnon,
  Yanghai Tsin, and Long Quan.
\newblock Neilf: Neural incident light field for physically-based material
  estimation.
\newblock In Shai Avidan, Gabriel~J. Brostow, Moustapha Ciss{\'{e}},
  Giovanni~Maria Farinella, and Tal Hassner, editors, {\em Computer Vision -
  {ECCV} 2022 - 17th European Conference, Tel Aviv, Israel, October 23-27,
  2022, Proceedings, Part {XXXI}}, volume 13691 of {\em Lecture Notes in
  Computer Science}, pages 700--716. Springer, 2022.

\bibitem{DBLP:conf/nips/YarivGKL21}
Lior Yariv, Jiatao Gu, Yoni Kasten, and Yaron Lipman.
\newblock Volume rendering of neural implicit surfaces.
\newblock In Marc'Aurelio Ranzato, Alina Beygelzimer, Yann~N. Dauphin, Percy
  Liang, and Jennifer~Wortman Vaughan, editors, {\em Advances in Neural
  Information Processing Systems 34: Annual Conference on Neural Information
  Processing Systems 2021, NeurIPS 2021, December 6-14, 2021, virtual}, pages
  4805--4815, 2021.

\bibitem{DBLP:conf/nips/YarivKMGABL20}
Lior Yariv, Yoni Kasten, Dror Moran, Meirav Galun, Matan Atzmon, Ronen Basri,
  and Yaron Lipman.
\newblock Multiview neural surface reconstruction by disentangling geometry and
  appearance.
\newblock In Hugo Larochelle, Marc'Aurelio Ranzato, Raia Hadsell,
  Maria{-}Florina Balcan, and Hsuan{-}Tien Lin, editors, {\em Advances in
  Neural Information Processing Systems 33: Annual Conference on Neural
  Information Processing Systems 2020, NeurIPS 2020, December 6-12, 2020,
  virtual}, 2020.

\bibitem{DBLP:conf/cvpr/YuYTK21}
Alex Yu, Vickie Ye, Matthew Tancik, and Angjoo Kanazawa.
\newblock pixelnerf: Neural radiance fields from one or few images.
\newblock In {\em {IEEE} Conference on Computer Vision and Pattern Recognition,
  {CVPR} 2021, virtual, June 19-25, 2021}, pages 4578--4587. Computer Vision
  Foundation / {IEEE}, 2021.

\bibitem{DBLP:journals/ijcv/YuC05}
Yizhou Yu and Johnny~T. Chang.
\newblock Shadow graphs and 3d texture reconstruction.
\newblock {\em Int. J. Comput. Vis.}, 62(1-2):35--60, 2005.

\bibitem{DBLP:conf/cvpr/ZhangLLS22}
Kai Zhang, Fujun Luan, Zhengqi Li, and Noah Snavely.
\newblock {IRON:} inverse rendering by optimizing neural sdfs and materials
  from photometric images.
\newblock In {\em {IEEE/CVF} Conference on Computer Vision and Pattern
  Recognition, {CVPR} 2022, New Orleans, LA, USA, June 18-24, 2022}, pages
  5555--5564. {IEEE}, 2022.

\bibitem{DBLP:conf/cvpr/ZhangLWBS21}
Kai Zhang, Fujun Luan, Qianqian Wang, Kavita Bala, and Noah Snavely.
\newblock Physg: Inverse rendering with spherical gaussians for physics-based
  material editing and relighting.
\newblock In {\em {IEEE} Conference on Computer Vision and Pattern Recognition,
  {CVPR} 2021, virtual, June 19-25, 2021}, pages 5453--5462. Computer Vision
  Foundation / {IEEE}, 2021.

\bibitem{DBLP:journals/tog/ZhangSDDFB21}
Xiuming Zhang, Pratul~P. Srinivasan, Boyang Deng, Paul~E. Debevec, William~T.
  Freeman, and Jonathan~T. Barron.
\newblock Nerfactor: neural factorization of shape and reflectance under an
  unknown illumination.
\newblock {\em {ACM} Trans. Graph.}, 40(6):237:1--237:18, 2021.

\bibitem{DBLP:conf/cvpr/ZhangSHFJZ22}
Yuanqing Zhang, Jiaming Sun, Xingyi He, Huan Fu, Rongfei Jia, and Xiaowei Zhou.
\newblock Modeling indirect illumination for inverse rendering.
\newblock In {\em {IEEE/CVF} Conference on Computer Vision and Pattern
  Recognition, {CVPR} 2022, New Orleans, LA, USA, June 18-24, 2022}, pages
  18622--18631. {IEEE}, 2022.

\end{thebibliography}
}

\appendix

\section{Relationship between camera and shadow ray supervision}

Ray supervision is the core of our method.
As the ray supervision is general for arbitrary rays, it leads to a dual relationship between camera ray supervision (\eg NeRF\cite{DBLP:conf/eccv/MildenhallSTBRN20}) and our method.
We list each method's components in \cref{tab:dual_nerf} to better illustrate their correspondences.
\begin{table*}[t]\centering
  \small
  \begin{tabular}{lll}\toprule
  &Camera ray supervision (NeRF) &Shadow ray supervision (Ours) \\\midrule
  Ray direction &View direction &Light direction \\
  Ray starting point &Camera location &Surface location \\
  Supervision label &Incoming radiance at the camera &Incoming radiance at the surface \\
  Particle-ray interactions &Absorption and emission &Absorption \\
  Capture setup &Multiple views &Multiple lights \\
  \bottomrule
  \end{tabular}
  \caption{Corresponding components in camera and shadow ray supervision.}\label{tab:dual_nerf}
  \end{table*}

\section{Additional implementation details}

\noindent {\bf Network architecture.} 
We adopt an 8-layer geometry MLP following \cite{DBLP:conf/nips/WangLLTKW21}.
When handling RGB inputs, we model another 4-layer material MLP.
We use Softplus for the geometry MLP and ReLU for the material MLP as activation. 
The hidden layers for both MLPs are 256 dimensional.
A 3D position with 6-frequency positional encoding is used as the input for the geometry MLP.
The geometry MLP outputs a signed distance and a 256-dimensional feature vector.
The feature vector is then concatenated with the 3D position and normal vector as the input for the material MLP.
The material MLP outputs a 3-channel diffuse albedo and 27 specular coefficients, with output activation by Softplus ($\beta=100$).
The specular coefficients are used to linearly combine nine spherical Gaussian bases with different shininess to produce a 3-channel specular color.
The diffuse and specular colors are represented in the linear color space.

\noindent {\bf Training.}
Our networks are trained using Adam\cite{DBLP:journals/corr/KingmaB14}, with the learning rate first linearly warmed up from 0 to $10^{-3}$ in the first 5k iterations and then cosine decayed to a minimum learning rate of $5\times 10^{-5}$.
The weight of the Eikonal loss is set to $0.01$, which we find a lower weight leads to more thin structures reconstructed.

\noindent {\bf Shadow ray sampling.} 
We place 80 uniform samples along the shadow ray and use the hierarchical sampling strategy in \cite{DBLP:conf/nips/WangLLTKW21} to sample another 64 points near the surface.
The far bound is determined by a scene bounding sphere.
The near bound is set to 0 so that detailed shadows by sample points near the starting surface can be modeled.
We are able to model these near sample points because the SDF-to-density formula (Eq. (3) in the main paper) is dependent on the ray and normal direction.
This property is suitable for modeling rays that start at the surface.
When the ray goes outward (the dot product between the ray direction and normal direction is greater than 0), we obtain zero densities at near sample points. 
Thus, the ray will not be incorrectly blocked by its starting surface.
When the ray goes inward, it will be appropriately occluded by the starting surface, generating attached shadows.

\begin{figure}[t]
  \centering
  \includegraphics[width=.98\linewidth]{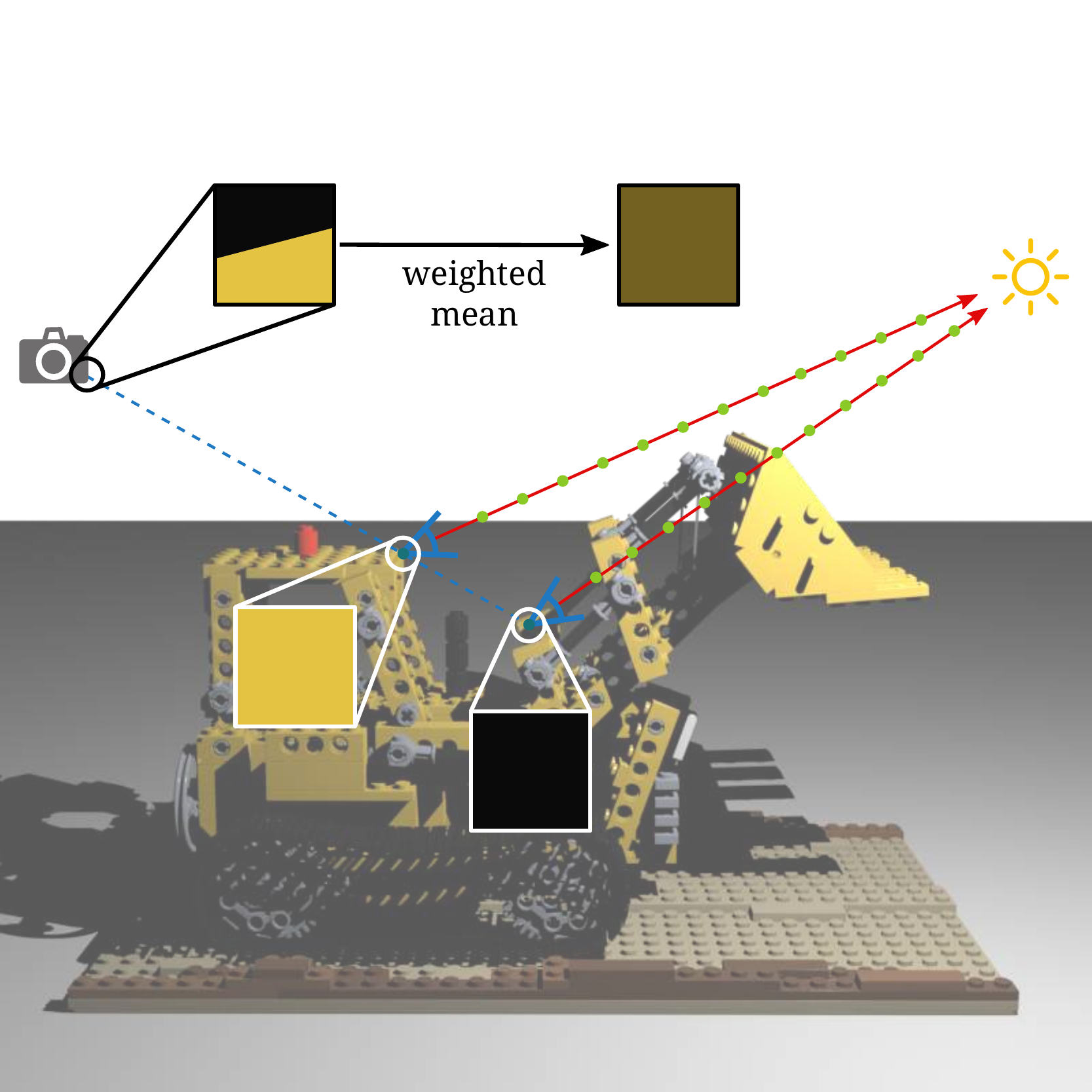}
  \caption{At a boundary pixel, we compute two shadow rays started at different depths and combine their results by weighted mean.}
  \label{fig:boundary_sample}
\end{figure}

\noindent {\bf Camera ray intersection.}
We use ray marching with 256 steps to locate the intersection between a camera ray and the SDF.
We then use a surface walk process in \cite{DBLP:conf/cvpr/ZhangLLS22} to locate the boundary points.
The surface walk process starts at the intersection points with a maximum of 16 steps.
In each step, a point moves along the surface with a step size of $2\times 10^{-3}$ until it reaches a boundary point whose surface normal direction is perpendicular to the camera ray direction.
The boundary point separates a pixel into two regions.
We locate the intersection points in the two sub-pixel regions using ray marching and compute the shadow rays started at each region respectively, as shown in \cref{fig:boundary_sample}.
The results of the shadow rays are combined by an area ratio proportional to each region. 
The area ratio is made differentiable by relating the area to the deformation of the boundary point.

Our setting differs from \cite{DBLP:conf/cvpr/ZhangLLS22} in that while they use edge sampling to refine an initial geometry, we are optimizing a geometry from scratch.
To accelerate convergence, we adopt a coarse-to-fine strategy that optimizes 100$\times$100 low-resolution images in the first 5k iterations and progressively upscales the images to the full 800$\times$800 resolution.
This strategy enlarges the pixel footprint, resulting in more boundary points to be considered in the early training iterations.

\section{Additional comparison results}

\subsection{Quantitative comparison on binary shadow inputs}

We evaluate two binary shadow datasets: A terrain-like dataset proposed by DeepShadow\cite{DBLP:journals/corr/abs-2203-15065} and a  non-terrain dataset proposed by us.
The results on the DeepShadow dataset are shown in \cref{tab:deepshadow_quant}, and the results on our dataset are shown in \cref{tab:shadow_quant}, respectively.
Our depth reconstruction outperforms DeepShadow on both terrain-like and non-terrain scenes.
Our normal reconstruction is better than DeepShadow on non-terrain scenes and comparable on terrain-like scenes.
\begin{table*}[!htp]\centering
  \footnotesize
  \begin{tabular}{lrrrrrrrrr}\toprule
  Method &Metric &Cactus &Rose &Bread &Sculptures &Surface &Relief &Avg \\\midrule
  DeepShadow &Depth L1$\downarrow$ &0.0091 &\textbf{0.0132} &0.0634 &0.0334 &0.0078 &0.0067 &0.0223 \\
  Ours &Depth L1$\downarrow$ &\textbf{0.0063} &0.0202 &\textbf{0.0256} &\textbf{0.0199} &\textbf{0.0036} &\textbf{0.0053} &\textbf{0.0135} \\\midrule
  DeepShadow &Normal MAE$\downarrow$ &20.79 &24.32 &\textbf{22.44} &26.66 &12.15 &\textbf{19.19} &20.93 \\
  Ours &Normal MAE$\downarrow$ &\textbf{20.02} &\textbf{18.35} &27.37 &\textbf{23.19} &\textbf{7.04} &22.13 &\textbf{19.68} \\
  \bottomrule
  \end{tabular}
  \caption{Quantitative comparison of reconstruction quality on the DeepShadow dataset.}\label{tab:deepshadow_quant}
  \end{table*}

  \begin{table*}[!htp]\centering
    \footnotesize
    \begin{tabular}{lrrrrrrrrrrr}\toprule
      Method &Metric &Chair &Drums &Ficus &Hotdog &Lego &Materials &Mic &Ship &Avg \\\midrule
      DeepShadow &Depth L1$\downarrow$ &0.7107 &0.1855 &1.6975 &0.0123 &0.4365 &0.0134 &0.8787 &0.0810 &0.5020 \\
      Ours &Depth L1$\downarrow$ &\textbf{0.0945} &\textbf{0.0532} &\textbf{1.1930} &\textbf{0.0054} &\textbf{0.0287} &\textbf{0.0119} &\textbf{0.0689} &\textbf{0.0408} &\textbf{0.1870} \\\midrule
      DeepShadow &Normal MAE$\downarrow$ &51.88 &18.98 &\textbf{25.48} &21.51 &38.42 &20.81 &31.87 &28.71 &29.71 \\
      Ours &Normal MAE$\downarrow$ &\textbf{18.08} &\textbf{13.27} &36.84 &\textbf{10.51} &\textbf{24.94} &\textbf{12.01} &\textbf{24.23} &\textbf{21.83} &\textbf{20.21} \\
      \bottomrule
      \end{tabular}      
    \caption{Quantitative comparison of reconstruction quality on our binary shadow dataset.}\label{tab:shadow_quant}
    \end{table*}

The normalized mean depth error (Depth nMZE) used in DeepShadow's paper is only suitable for terrain-like scenes.
Therefore, we propose to compute depth error by aligning the depth map to the ground truth using ICP (denoted as Depth L1).
For completeness, we also show quantitative results on the DeepShadow dataset using normalized mean depth error in \cref{tab:deepshadow_nmze}.
We report DeepShadow's results from their publicly available code, which are slightly better than their paper results.

\begin{table*}[!htp]\centering
  \footnotesize
  \begin{tabular}{lrrrrrrrrr}\toprule
    Method &Metric &Cactus &Rose &Bread &Sculptures &Surface &Relief &Avg \\\midrule
    DeepShadow &Depth nMZE$\downarrow$ &0.1001 &0.0760 &0.1166 &0.1779 &0.0952 &\textbf{0.1424} &0.1180 \\
    Ours &Depth nMZE$\downarrow$ &\textbf{0.0392} &\textbf{0.0709} &\textbf{0.1001} &\textbf{0.0678} &\textbf{0.0381} &0.1427 &\textbf{0.0765} \\
    \bottomrule
    \end{tabular}
  \caption{Quantitative comparison on the DeepShadow dataset using normalized mean depth error.}\label{tab:deepshadow_nmze}
  \end{table*}

\begin{figure}[ht]
  \centering
  \includeinkscape[width=.98\linewidth]{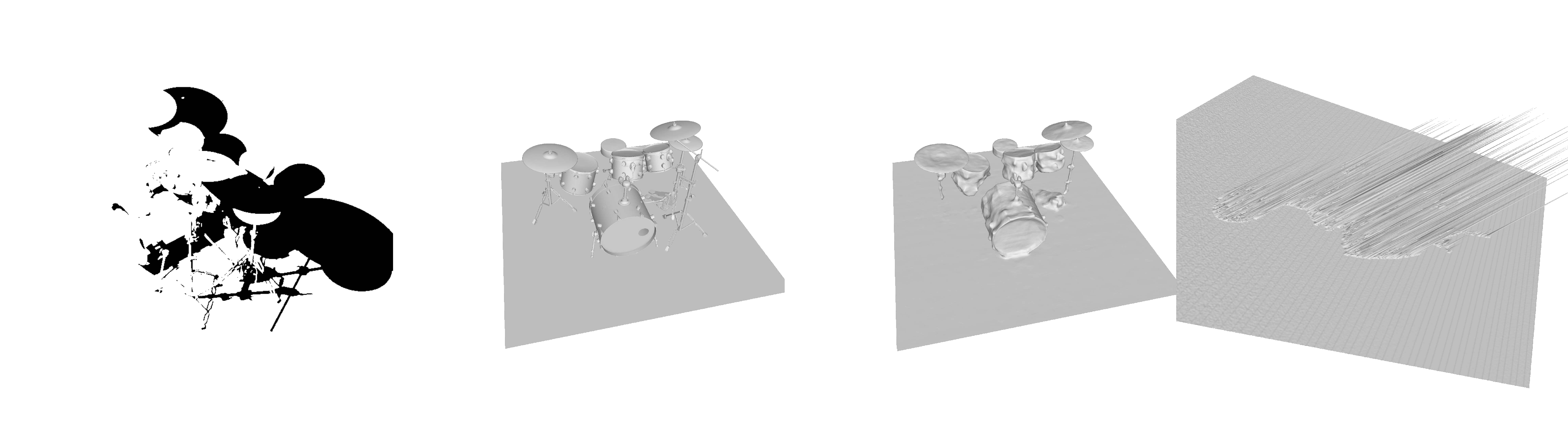}
  \includeinkscape[width=.98\linewidth]{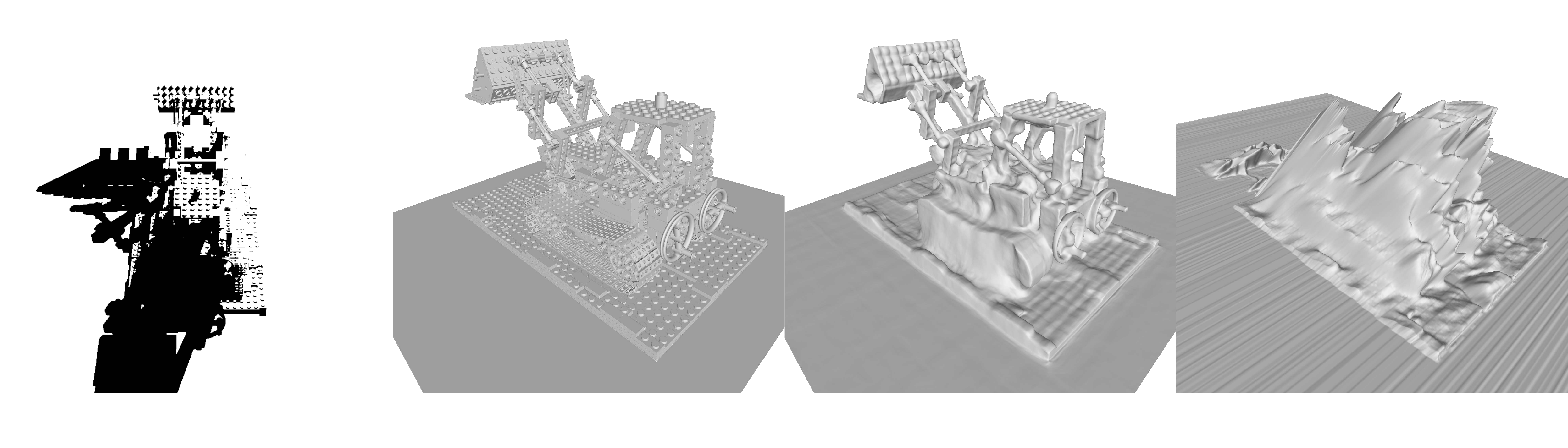}
  \caption{Qualitative comparison on our side-view binary shadow dataset.}
  \label{fig:deepshadow_side}
\end{figure}

\subsection{Qualitative comparison on our side-view binary shadow inputs}
We mainly conduct comparisons on our binary shadow dataset using a vertical-down viewpoint because previous works that adopt a depth map representation work better at a vertical-down camera.
For completeness, we provide qualitative comparison results on our side-view binary shadow dataset in \cref{fig:deepshadow_side}.

\subsection{Quantitative comparison on RGB inputs}

We show the quantitative results of SDPS-Net\cite{DBLP:conf/cvpr/ChenHSMW19}, Li et al. \cite{DBLP:conf/cvpr/LiL22} and our method on our RGB dataset in \cref{tab:rgb_quant}.
We achieve the lowest depth and normal reconstruction error.

\begin{table*}[!htp]\centering
  \footnotesize
  \begin{tabular}{lrrrrrrrrrrr}\toprule
    Method &Metrics &Chair &Drums &Ficus &Hotdog &Lego &Materials &Mic &Ship &Avg \\\midrule
    SDPS-Net &Depth L1$\downarrow$ &1.2627 &0.8706 &1.9185 &0.5964 &0.7254 &0.1700 &1.3678 &0.4190 &0.9163 \\
    Li et al. &Depth L1$\downarrow$ &1.2285 &0.9467 &1.8904 &0.1372 &0.6376 &0.8242 &1.2676 &\textbf{0.1027} &0.8794 \\
    Ours &Depth L1$\downarrow$ &\textbf{0.0090} &\textbf{0.0383} &\textbf{0.7959} &\textbf{0.0145} &\textbf{0.0316} &\textbf{0.0057} &\textbf{0.0419} &0.1360 &\textbf{0.1341} \\\midrule
    SDPS-Net &Normal MAE$\downarrow$ &31.90 &31.59 &55.65 &42.10 &39.00 &31.11 &34.92 &45.21 &38.94 \\
    Li et al. &Normal MAE$\downarrow$ &14.72 &25.93 &\textbf{34.60} &9.31 &21.77 &43.49 &25.68 &13.34 &23.61 \\
    Ours &Normal MAE$\downarrow$ &\textbf{7.65} &\textbf{17.09} &37.73 &\textbf{6.70} &\textbf{17.87} &\textbf{9.21} &\textbf{11.95} &\textbf{12.02} &\textbf{15.03} \\
    \bottomrule
    \end{tabular}
  \caption{Quantitative comparison of reconstruction quality on our RGB dataset.}\label{tab:rgb_quant}
  \end{table*}

\section{Discussion on the handling of ground}

\subsection{Results on non-planar grounds}
\begin{figure*}[ht]
  \centering
  \begin{minipage}[t]{.49\textwidth}
    \includeinkscape[width=.98\linewidth]{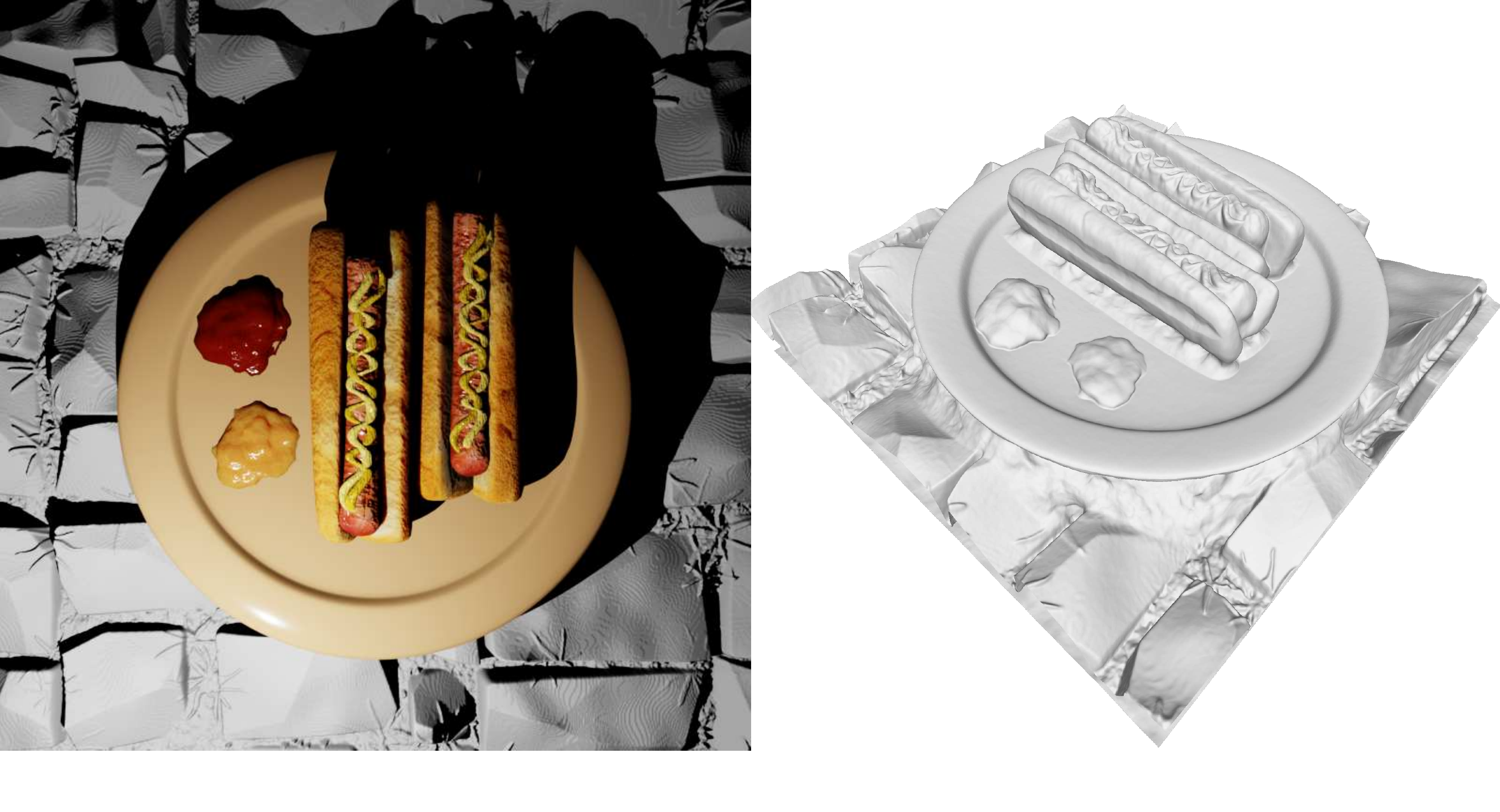}
    \includeinkscape[width=.98\linewidth]{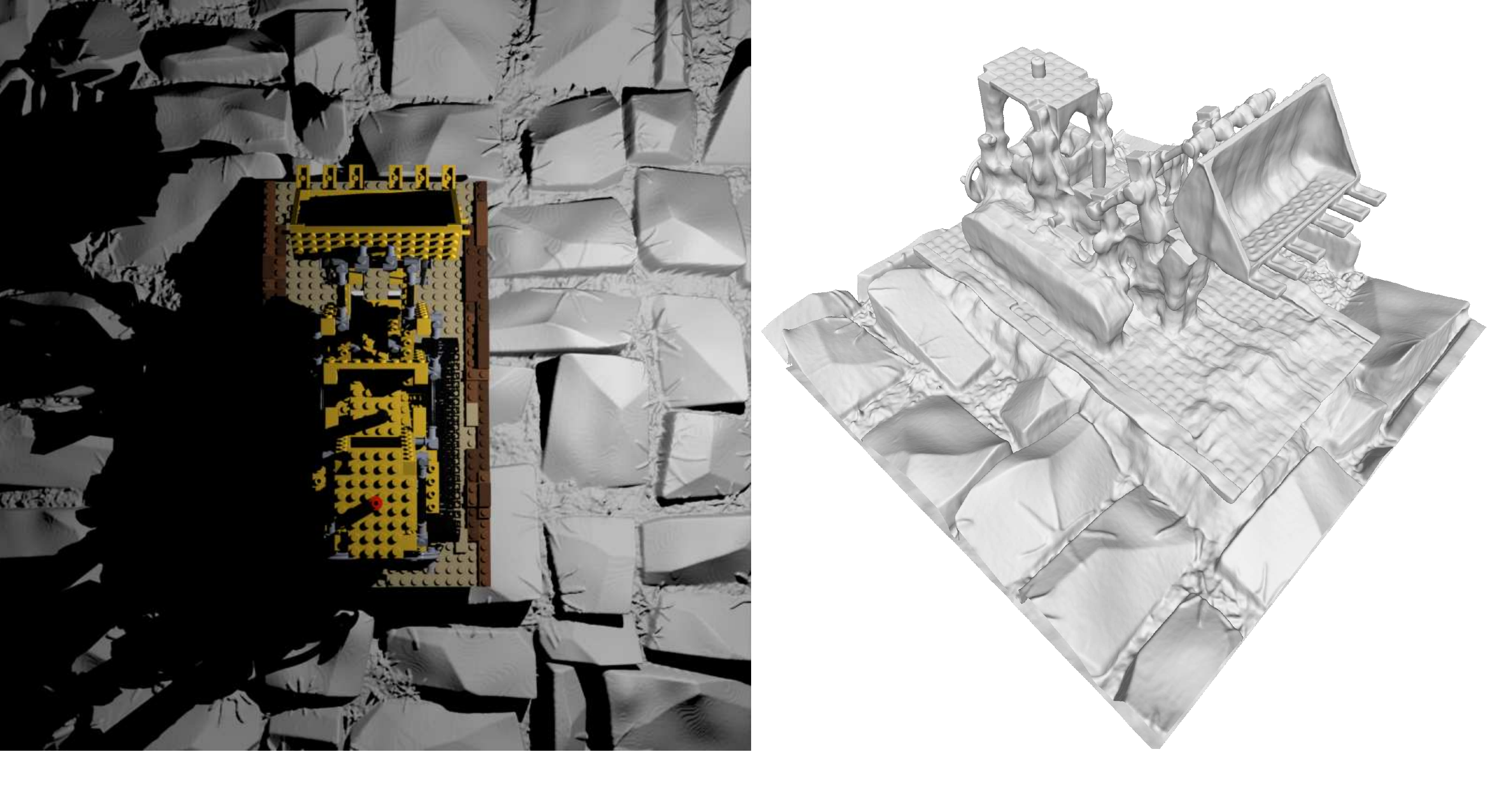}
  \end{minipage}
  \begin{minipage}[t]{.49\textwidth}
    \includeinkscape[width=.98\linewidth]{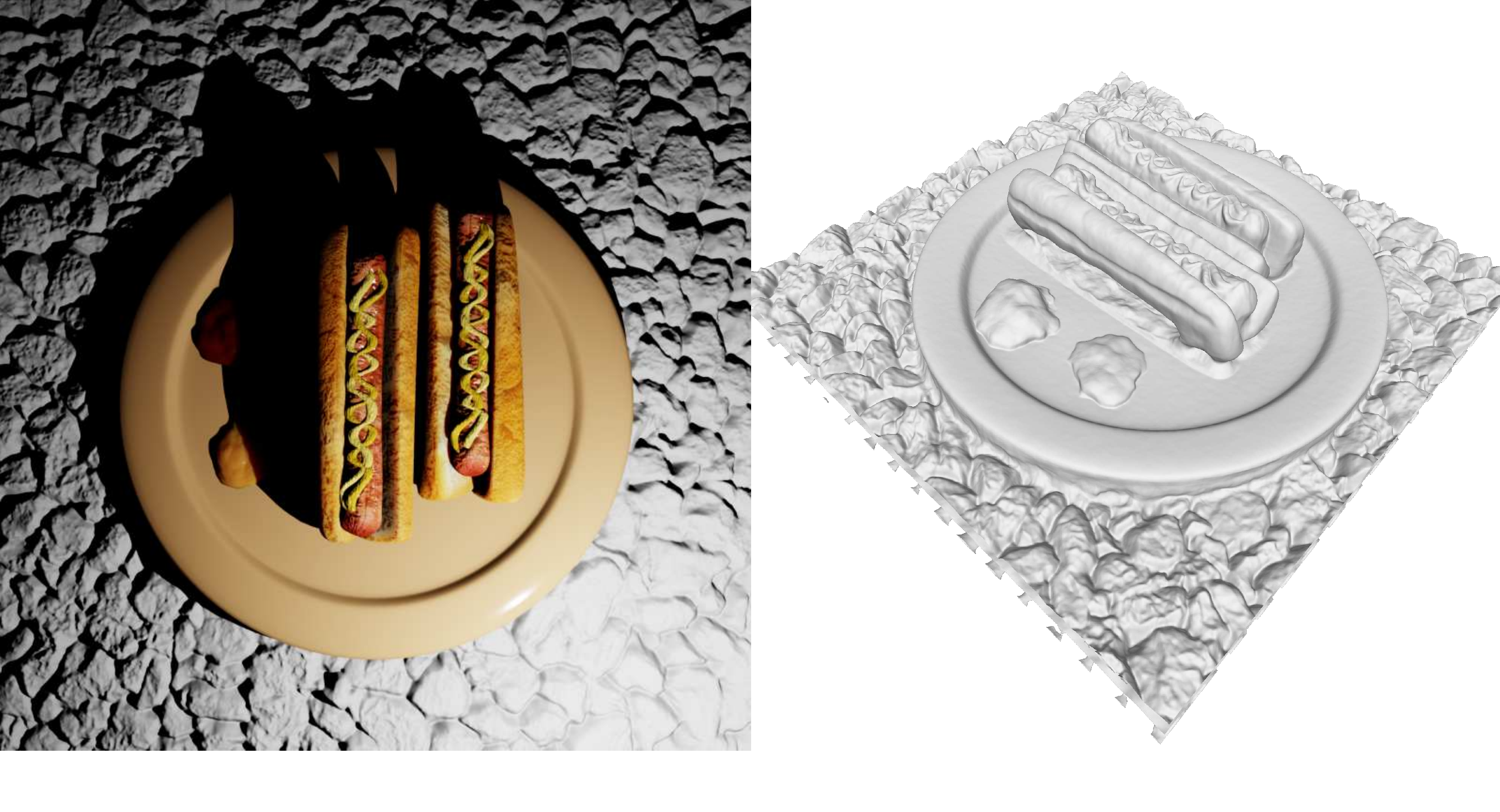}
    \includeinkscape[width=.98\linewidth]{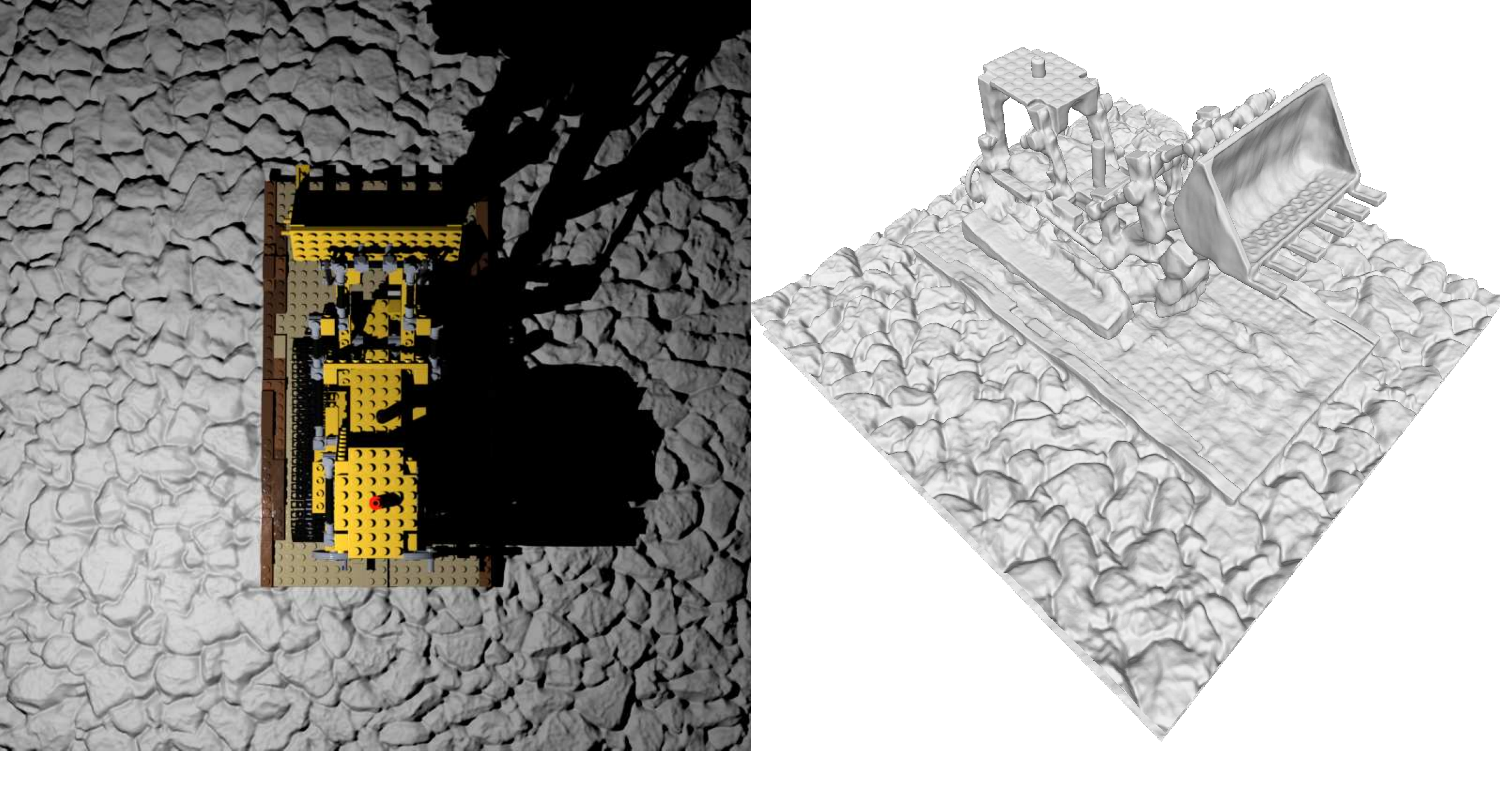}
  \end{minipage}
  \caption{Results in the presence of bumpy grounds.}
  \label{fig:complex_ground}
\end{figure*}

Given single-view images, the scale of the reconstructed scene is unconstrained.
One possible way to resolve scale ambiguities is to calibrate the ground position, which is adopted in the evaluation of our method.
We mainly evaluate planar grounds because they are common in real-world indoor setups and can easily calibrate by a checkerboard.
However, our method is not inherently limited to planar grounds.
When the ground is non-planar, we require that the depth map of the ground is known.
We initialize the ground surface by regularizing the SDF at the ground to be 0.
As shown in \cref{fig:complex_ground}, our method successfully reconstructs the object shapes in the presence of bumpy grounds.

\subsection{Comparison between known and unknown grounds}

\begin{figure*}[ht]
  \centering
  \includeinkscape[width=.98\linewidth]{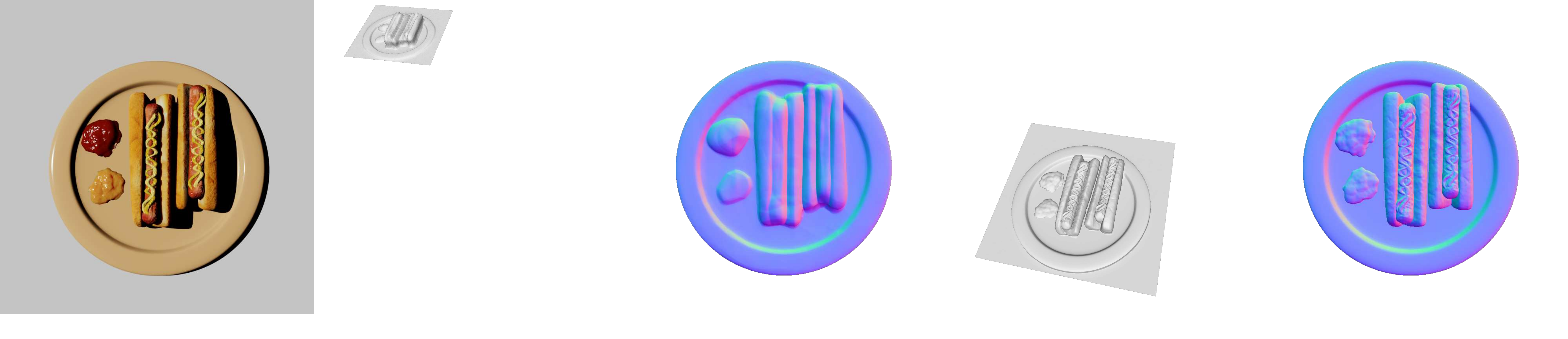}
  \vspace{0.5cm}
  \includeinkscape[width=.98\linewidth]{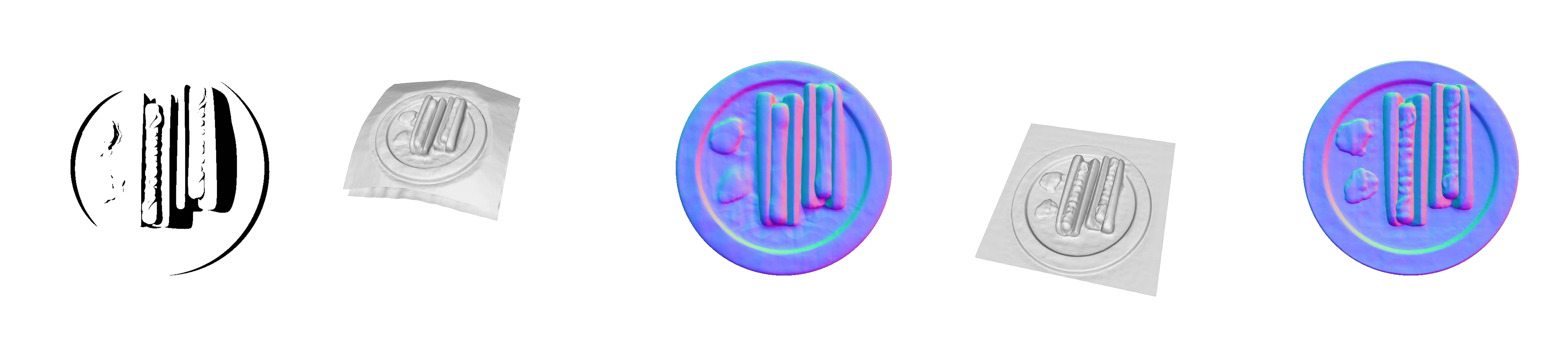}
  \vspace{0.5cm}
  \includeinkscape[width=.98\linewidth]{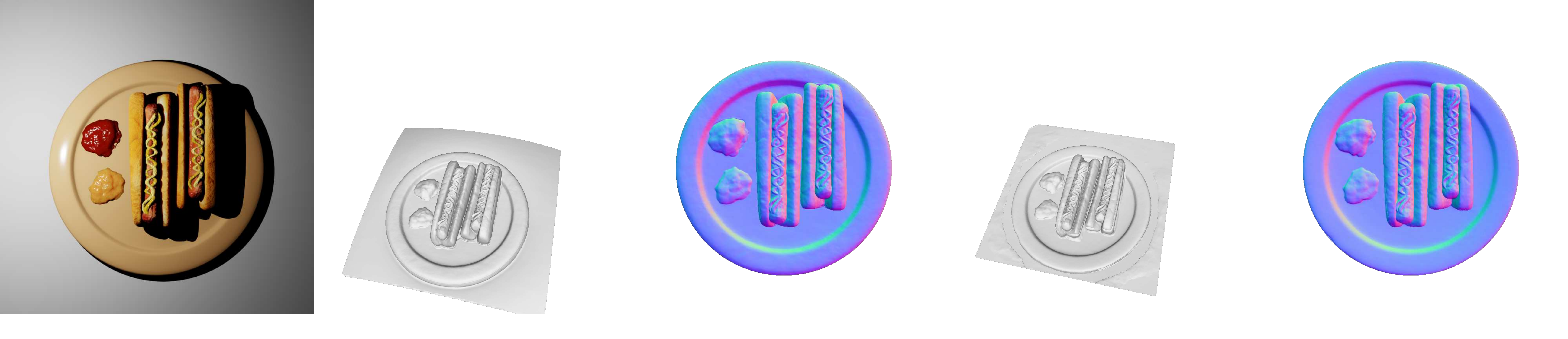}
  \vspace{0.5cm}
  \includeinkscape[width=.98\linewidth]{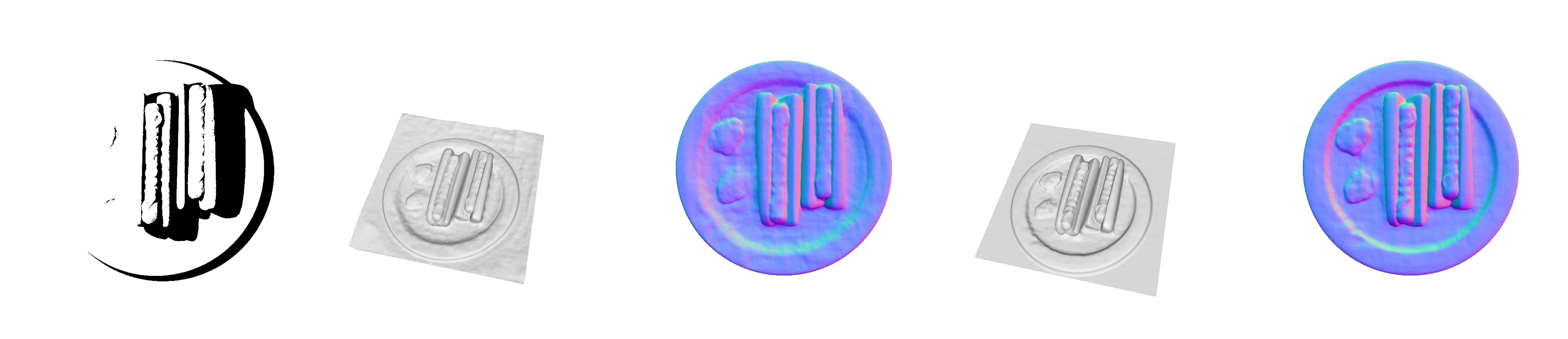}
  \vspace{0.5cm}
  \includeinkscape[width=.98\linewidth]{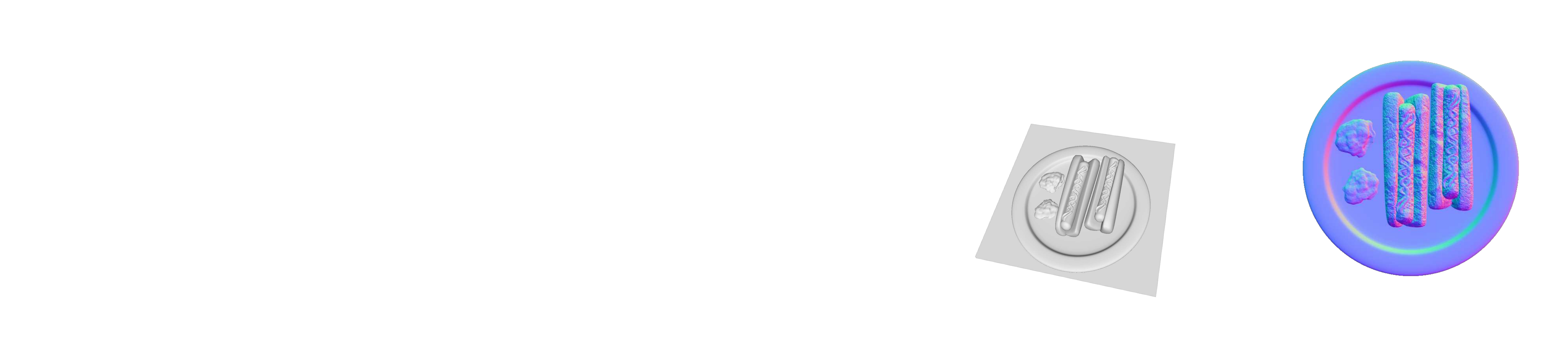}
  \caption{Comparison between known and unknown grounds.}
  \label{fig:known_ground}
\end{figure*}

To investigate the effect of the ground, we compare results with known and unknown grounds under different input types.
As shown in \cref{fig:known_ground}, our method still achieves reasonable reconstruction when the ground is unknown, but the reconstruction exhibits a scale drift, especially when using directional light inputs.
When the scale of the reconstruction deviates, its quality also decreases, possibly because it only occupies a small portion of the scene bounding sphere.
Therefore, we choose to calibrate the ground in the evaluation to obtain scale-accurate reconstruction under arbitrary input types.

\begin{figure}[ht]
  \centering
  \includeinkscape[width=.98\linewidth]{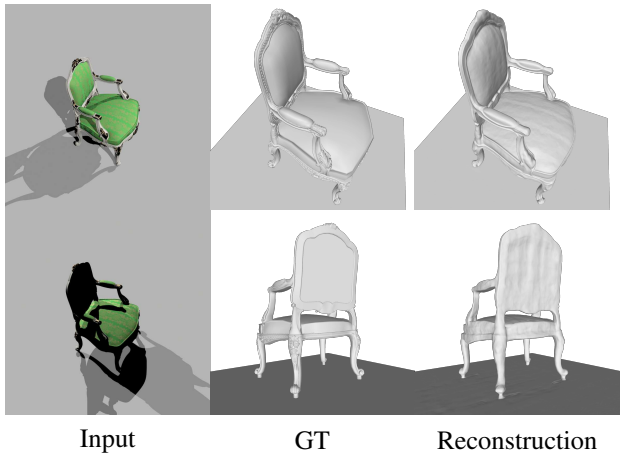}
  \caption{Results on the scene illuminated by two lights.}
  \label{fig:2light}
\end{figure}

\section{Additional evaluation}

\subsection{Analysis on the number of input images}

\begin{figure*}[t]
  \centering
  \includeinkscape[width=.98\linewidth]{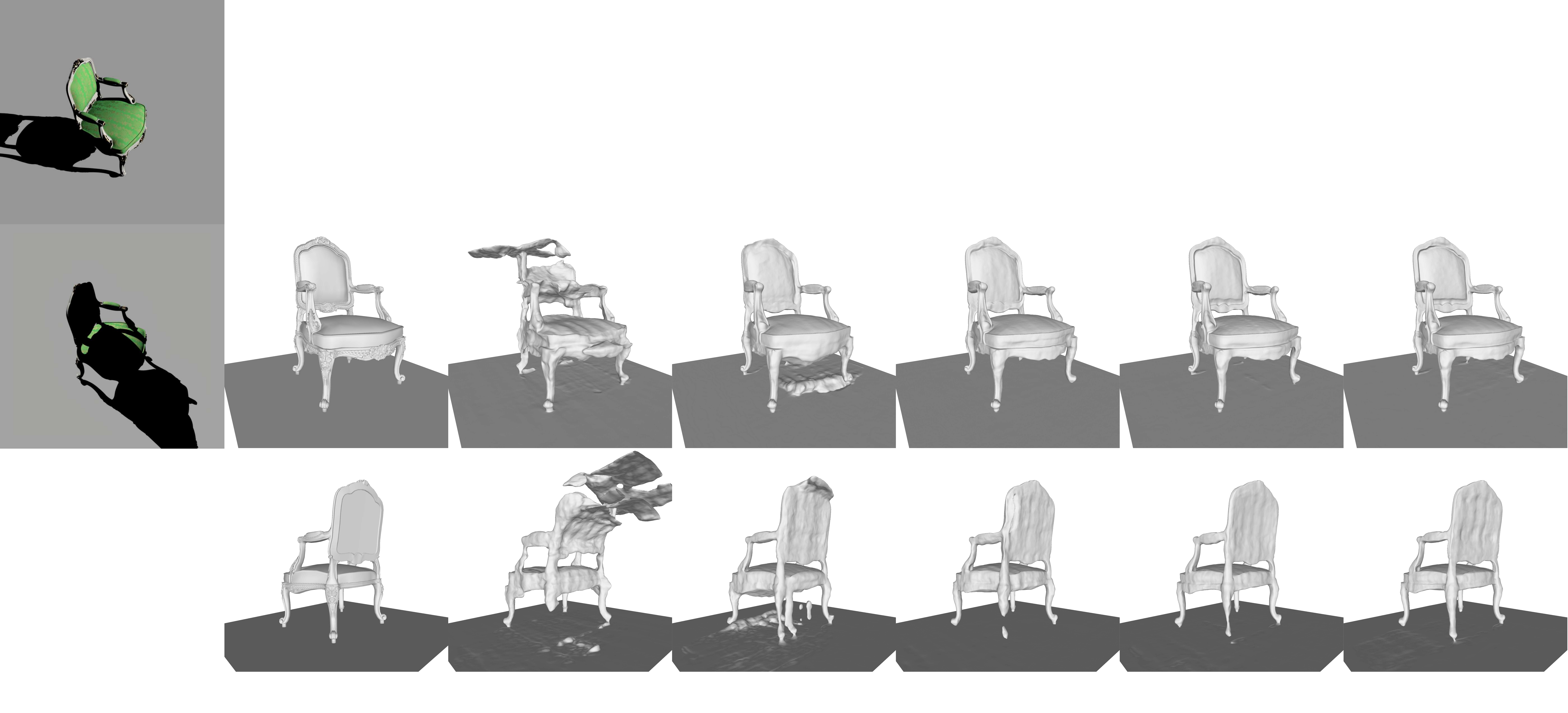}
  \caption{Analysis on different numbers of input images.}
  \label{fig:eval_number}
\end{figure*}

To investigate our method's robustness, we evaluate it on the {\it Chair} scene using different numbers of input images.
As shown in \cref{fig:eval_number} and \cref{tab:eval_number}, our method can reconstruct reasonable geometry under five input images. When the input image number increases, the reconstructed structures become more accurate.
In general, our method is robust to the number of input images.

\begin{table}[t]\centering
  \small
  \begin{tabular}{lrrr}\toprule
  Number of images &Depth L1$\downarrow$ &Normal MAE$\downarrow$ \\\midrule
  3 &0.1427 &28.03 \\
  5 &0.0216 &10.52 \\
  10 &0.0189 &8.88 \\
  20 &0.0127 &7.59 \\
  50 &\textbf{0.0074} &\textbf{7.01} \\
  \bottomrule
  \end{tabular}
  \caption{Reconstruction quality using different numbers of input images.}\label{tab:eval_number}
\end{table}

\subsection{Effect of foreground and background shadows in reconstruction}

\begin{figure*}[ht]
  \centering
  \includeinkscape[width=.98\linewidth]{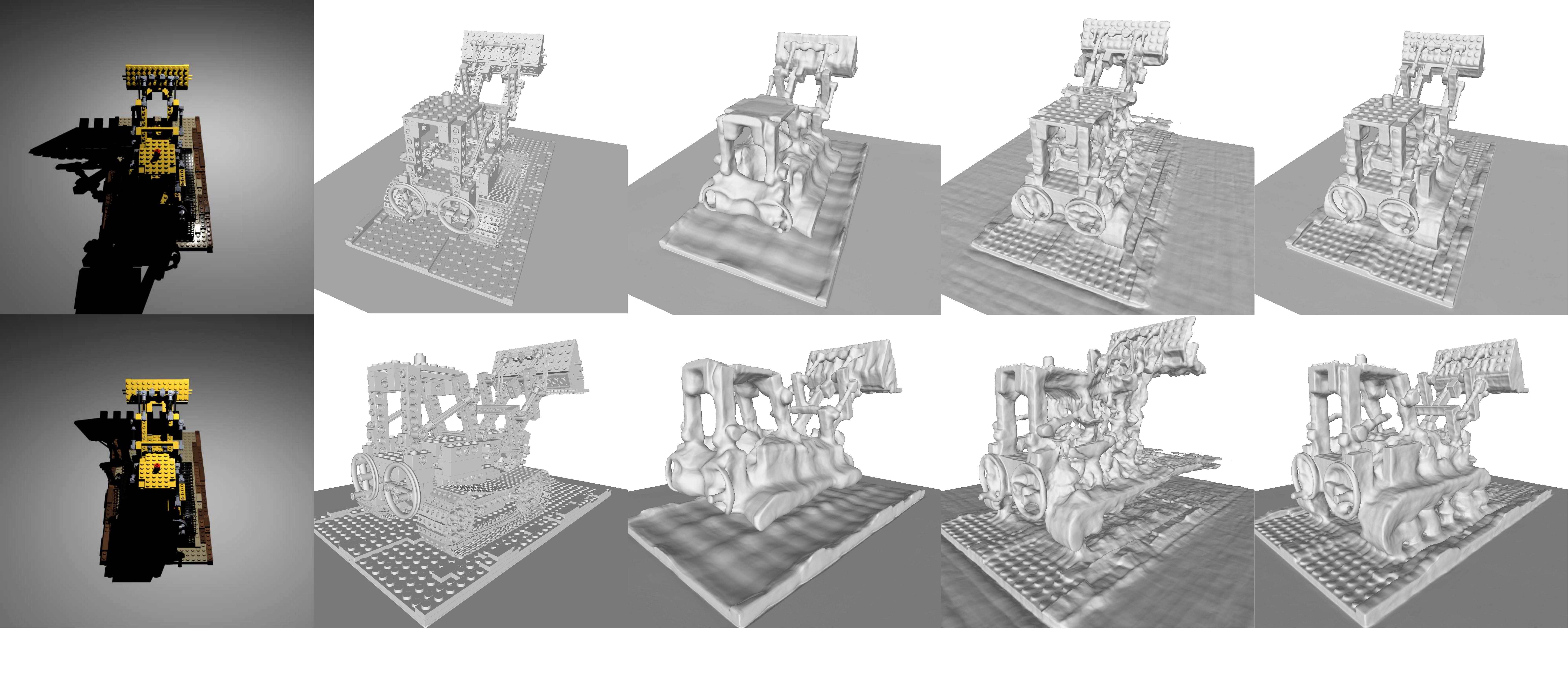}
  \caption{Comparison of shape reconstruction when supervising only background or foreground shadows.}
  \label{fig:fore_back}
\end{figure*}

To investigate how the supervision of foreground and background shadows affects shape reconstruction, we compare our method on the {\it Lego} scene with two variants that only supervise the background or foreground shadows.
As shown in \cref{fig:fore_back}, when we only supervise shadows cast on the ground, we cannot reconstruct detailed structures on the top of the bulldozer.
The middle part is also missing, as it mainly casts shadows on the object itself.
When we only supervise foreground shadows, we can reconstruct the detailed structures, but the reconstructed bulldozer shovel is at an incorrect depth.
As shown in \cref{tab:fore_back}, our method achieves the lowest reconstruction error when supervising foreground and background shadows.
The two parts of shadows are indispensable in accurate shape reconstruction.

\begin{table}[!htp]\centering
  \small
  \begin{tabular}{lrrr}\toprule
  &Depth L1$\downarrow$ &Normal MAE$\downarrow$ \\\midrule
  Back only &0.05827 &29.93 \\
  Fore only &0.13569 &23.94 \\
  Ours &\textbf{0.02955} &\textbf{19.59} \\
  \bottomrule
  \end{tabular}
  \caption{Reconstruction quality when supervising only background or foreground shadows.}\label{tab:fore_back}
  \end{table}

\subsection{Results on scene illuminated by two lights}

We mainly evaluate our method illuminated by one known light.
However, our method can be extended to handle multiple known lights.
As shown in \cref{fig:2light}, by supervising the sum of the incoming radiance of two lights, our method can still reconstruct a complete 3D shape of the chair.

\section{Applications}

\begin{figure*}[ht]
  \centering
  \includeinkscape[width=.98\linewidth]{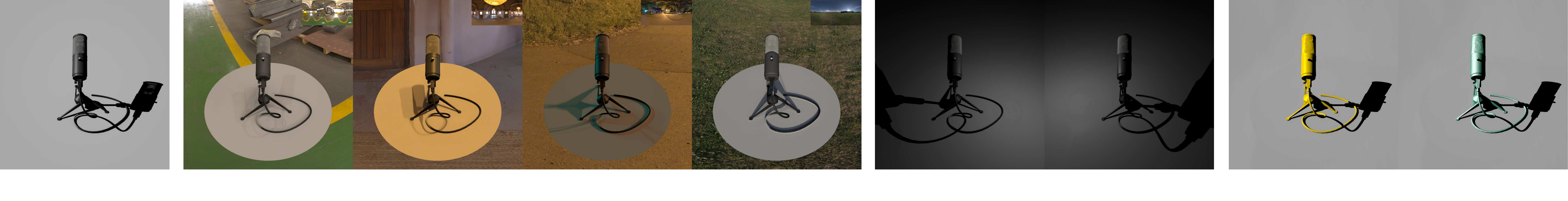}
  \includeinkscape[width=.98\linewidth]{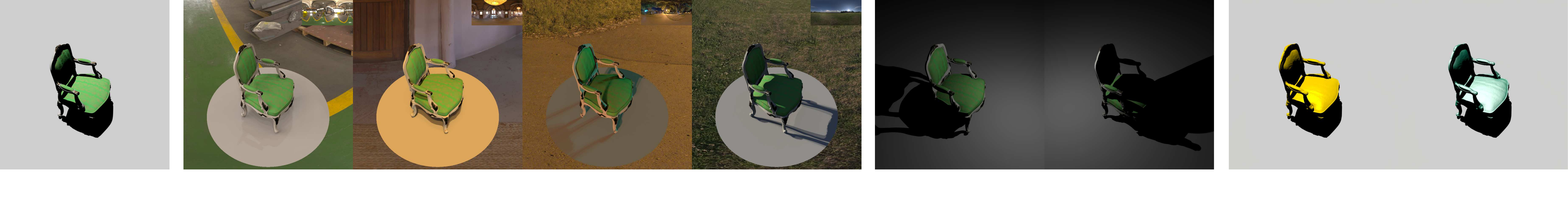}
  \includeinkscape[width=.98\linewidth]{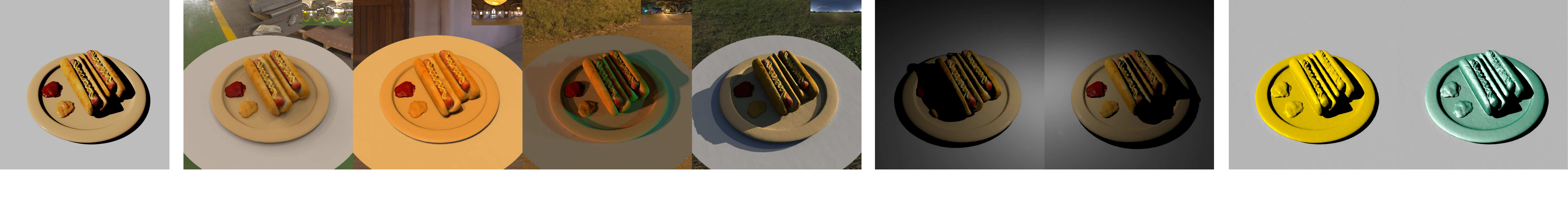}
  \caption{Applications.}
  \label{fig:editing_app}
\end{figure*}

Our method can reconstruct shapes and materials from single-view RGB images.
Therefore, it supports multiple applications, such as relighting using a point light or an environment map and material editing.
In \cref{fig:editing_app}, we show that our method generates plausible results in these applications.
Please also see the supplementary video for more results.

\section{Discussion on surface locating method}

\begin{figure}[h]
  \centering
  \includeinkscape[width=\linewidth]{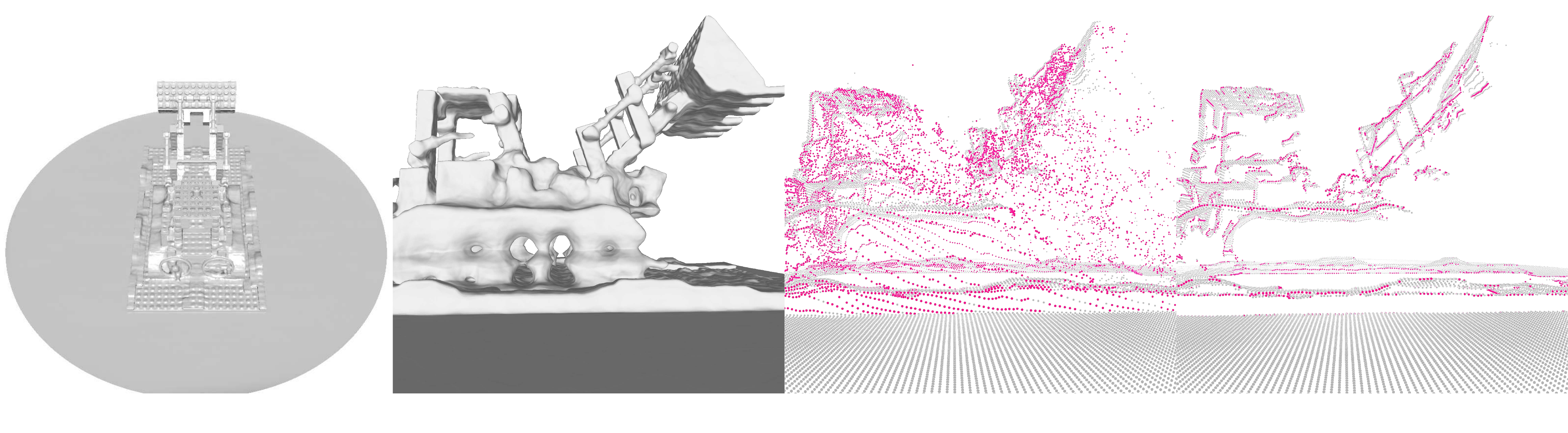}
  \caption{Visualized intersections of the {\it same} SDF (column 2) using the viewpoint in column 1. Boundaries are shown in magenta.}
  \label{fig:exp_depth_map}
\end{figure}
\begin{figure}[h]
  \centering
  \includeinkscape[width=\linewidth]{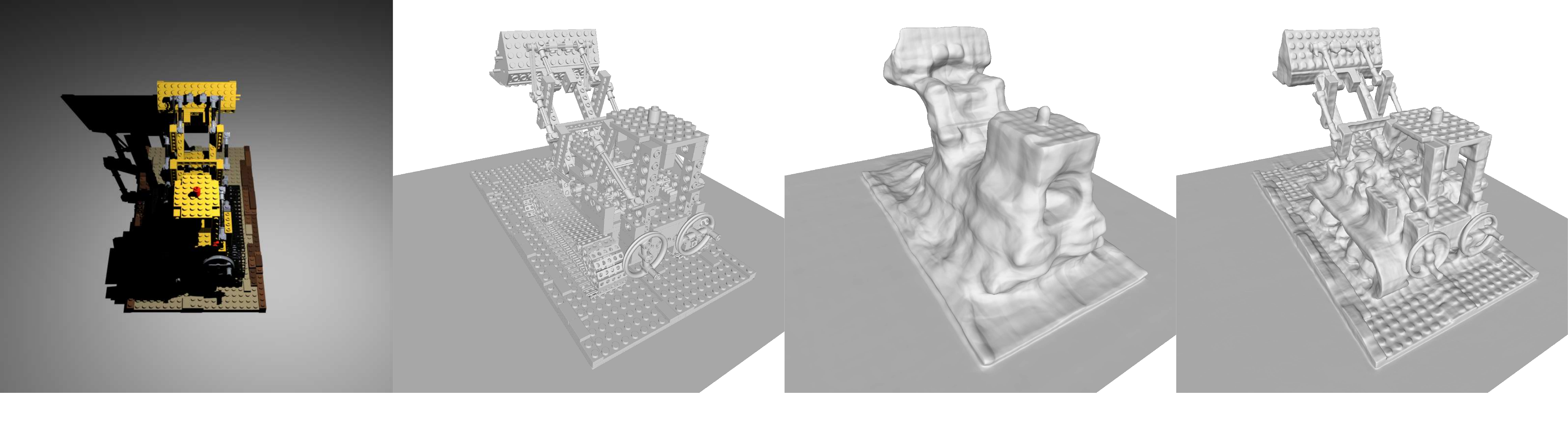}
  \caption{Qualitative comparison of reconstructed shape between ``expected depth'' and our method.}
  \label{fig:exp_depth_compare}
\end{figure}

We use NeuS-like volume rendering for shadow rays due to its wider basin of convergence \cite{10.1145/3306346.3323020}, which helps discover better reconstructions. 
However, for camera rays, straightforward NeuS-like volumetric sampling is impractically complex because each sample is costly and the sample count is too large. 
An alternative method to our proposed surface intersection is presented in \cite{DBLP:journals/tog/ZhangSDDFB21}, which computes expected terminated depth by weighting depth samples by volume densities. 
Both ``expected depth'' \cite{DBLP:journals/tog/ZhangSDDFB21} and our method are differentiable and reduce the sample count. 
However, we initially tried ``expected depth'' in early experiments and found that it computes incorrect ``averaged'' intersections at surface boundaries (\cref{fig:exp_depth_map} column 3). 
This greatly hindered optimization, as shown in the qualitative comparison in \cref{fig:exp_depth_compare}. 
By incorporating implicit differentiation \cite{DBLP:conf/nips/YarivKMGABL20} with edge sampling \cite{DBLP:conf/cvpr/ZhangLLS22}, our framework computes fully differentiable, correct intersections with a reasonable sample count (\cref{fig:exp_depth_map} column 4).

\section{Synthetic dataset examples}
\begin{figure*}[ht]
  \centering
  \includeinkscape[width=.98\linewidth]{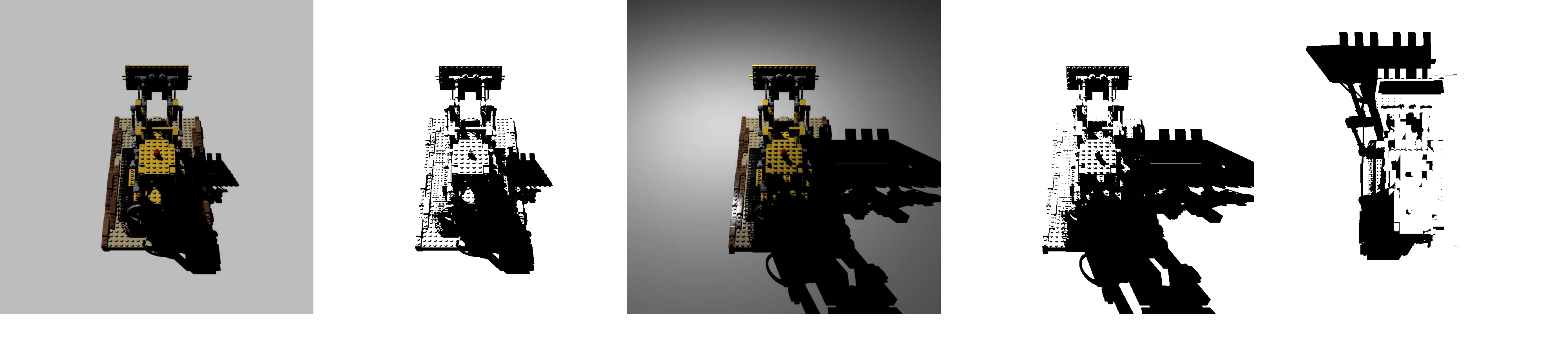}
  \includeinkscape[width=.98\linewidth]{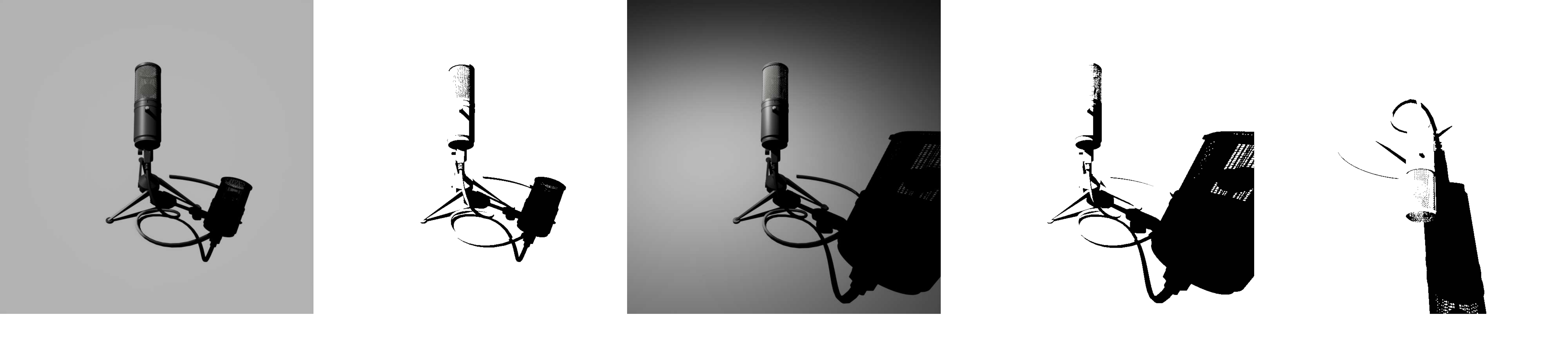}
  \caption{Example data from our synthetic dataset.}
  \label{fig:dataset_type}
\end{figure*}

In \cref{fig:dataset_type}, we show different data types from our synthetic dataset.

\section{Real dataset examples}

\begin{figure*}[ht]
  \centering
  \includeinkscape[width=.98\linewidth]{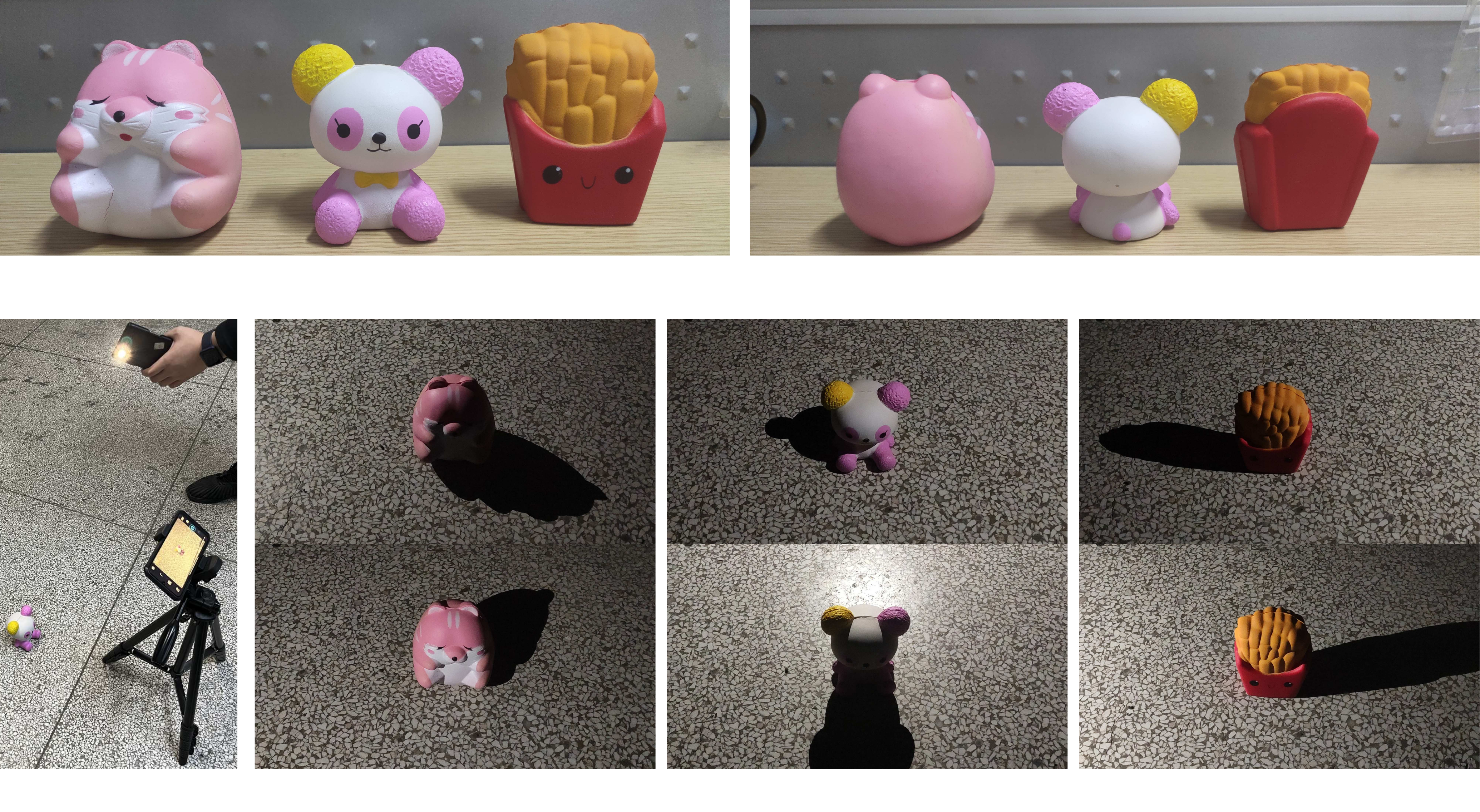}
  \caption{More details of our real dataset.}
  \label{fig:real_dataset}
\end{figure*}

In \cref{fig:real_dataset}, we show the objects, capture setup, and example images from our real dataset.

\section{Social impact}
As our method targets shape reconstruction from single-view inputs, it could be extended to be misused for improper surveillance.
In particular, 3D shapes can be reconstructed by exploiting shadows on the visible surface, revealing scenes beyond the camera's line of sight.

\end{document}